%% file: main.tex
\documentclass{article}

\usepackage{hyperref}

\usepackage[accepted]{icml_template/icml2024}

\usepackage{amsmath}
\usepackage{amssymb}
\usepackage{mathtools}
\usepackage{amsthm}
\usepackage{textcomp}

\usepackage[capitalize,noabbrev]{cleveref}

\theoremstyle{plain}
\newtheorem{theorem}{Theorem}[section]

\theoremstyle{definition}
\newtheorem{definition}[theorem]{Definition}

\icmltitlerunning{Incremental Fourier Neural Operator}

\usepackage{multirow}
\input{packages}

\input{macros}

\input{math_command}

\usepackage{graphicx}
\usepackage{pifont}
\usepackage{multicol}
\usepackage{tablefootnote}
\usepackage[para,online,flushleft]{threeparttable}

\usepackage{setspace}

\usepackage{algorithm}

\usepackage[many]{tcolorbox}

\usepackage{subcaption}

\usepackage{letltxmacro}

\usepackage{siunitx}
\begin{document}

\input{sections/0-title}

\input{sections/abs.tex}

\input{sections/intro.tex}
\input{sections/related_works.tex}
\input{sections/FNO/model_intro.tex}

\input{sections/importance.tex}

\input{sections/motivations.tex}

\input{sections/experiments.tex}

\input{sections/conclusion.tex}

\bibliography{zotero,custom}
\bibliographystyle{icml_template/icml2024}

\newpage
\clearpage
\onecolumn
\appendix
\input{sections/appendix/discussions.tex}
\input{sections/appendix/additional_results.tex}

\input{sections/appendix/rebuttal_results}

\end{document}

%% file: packages.tex
\usepackage{xcolor}
\usepackage{fancyhdr,graphicx}
\usepackage{amssymb,amsthm,amsmath}
\usepackage{booktabs,array} %
\usepackage{thmtools}
\usepackage{thm-restate}
\usepackage[title]{appendix}
\usepackage{tikz}
\usepackage{resizegather}
\usepackage{wrapfig}
\usepackage{enumitem,kantlipsum}
\usepackage{mathtools}
\usepackage{nccmath}

%% file: macros.tex
 \usepackage{xspace}

\usepackage{dsfont}
\usepackage{stmaryrd}
\usepackage{comment}

\renewcommand{\phi}{\varphi}

\newcommand{\Complex}{\mathbb{C}}

%% file: math_command.tex
\usepackage{amsmath,amsfonts,bm}

\def\eqref#1{equation~\ref{#1}}

\def\1{\bm{1}}

\DeclareMathAlphabet{\mathsfit}{\encodingdefault}{\sfdefault}{m}{sl}
\SetMathAlphabet{\mathsfit}{bold}{\encodingdefault}{\sfdefault}{bx}{n}

%% file: sections/0-title.tex
\twocolumn[
\icmltitle{Incremental Spatial and Spectral Learning of Neural Operators \\ for Solving Large-Scale PDEs}

\icmlsetsymbol{equal}{*}

\begin{icmlauthorlist}
\icmlauthor{Robert Joseph George}{equal,caltech}
\icmlauthor{Jiawei Zhao}{equal,caltech}
\icmlauthor{Jean Kossaifi}{nvidia}
\icmlauthor{Zongyi Li}{caltech}
\icmlauthor{Anima Anandkumar}{caltech}
\end{icmlauthorlist}

\icmlaffiliation{caltech}{California Institute of Technology}
\icmlaffiliation{nvidia}{NVIDIA}

\icmlcorrespondingauthor{Robert Joseph George}{rgeorge@caltech.edu}
\icmlcorrespondingauthor{Jiawei Zhao}{jiawei@caltech.edu}

\icmlkeywords{Machine Learning, ICML}

\vskip 0.3in
]

\printAffiliationsAndNotice{\icmlEqualContribution} %

%% file: sections/abs.tex
\begin{abstract}
Fourier Neural Operators (FNO) offer a principled approach to solving challenging partial differential equations (PDE) such as turbulent flows.
At the core of FNO is a spectral layer that leverages a discretization-convergent representation in the Fourier domain, and learns weights over a fixed set of frequencies.
However, training FNO presents two significant challenges, particularly in large-scale, high-resolution applications:
(i) Computing Fourier transform on high-resolution inputs is computationally intensive but necessary since fine-scale details are needed for solving many PDEs, such as fluid flows,
(ii) selecting the relevant set of frequencies in the spectral layers is challenging, and too many modes can lead to overfitting, while too few can lead to underfitting.
To address these issues, we introduce the \emph{Incremental Fourier Neural Operator} (iFNO), which progressively increases both the number of frequency modes used by the model as well as the resolution of the training data.
We empirically show that iFNO reduces total training time while maintaining or improving generalization performance across various datasets. Our method demonstrates a 10\% lower testing error, using 20\% fewer frequency modes compared to the existing Fourier Neural Operator, while also achieving a 30\% faster training.
\end{abstract}

%% file: sections/intro.tex
\section{Introduction}

Recently, deep learning has shown promise in solving partial differential equations (PDEs) significantly faster than traditional numerical methods in many domains. Among them, 
\emph{Fourier neural operator} (FNO), proposed by \citet{li_fourier_2021}, is a family of neural operators that achieves state-of-the-art performance in solving PDEs, in applications modeling turbulent flows, weather forecasting \cite{pathak2022fourcastnet}, material plasticity \cite{liu2022learning}, and carbon dioxide storage \cite{wen2022u}. FNO learns the solution operator of PDE that maps the input (initial and boundary conditions) to the output (solution functions).

Unlike conventional neural networks, FNO possesses a property termed \emph{discretization-convergence}, meaning it can output predictions at different resolutions, and those predictions converge to a unique solution upon mesh refinement.  The discretization-convergence property relies on kernel integration, which, in the case of FNO, is realized using the Fourier transform. A series of such spectral layers, along with normalization layers and non-linear activations, compose the core of the FNO architecture. For many PDE systems, like fluid flows, there is a decay of energy with frequency, i.e., low-frequency components usually have larger magnitudes than high-frequency ones.
Therefore, as an inductive bias and a regularization procedure to avoid overfitting, FNO consists of a frequency truncation function $T_K$ in each layer that only allows the lowest $K$ Fourier modes to propagate to the next layer, and the other modes are zeroed out, as shown in Figure~\ref{fig:ifno_framework}.

\input{figures/wrap_ifno_framework.tex}

It is thus crucial to select the correct number of frequency modes $K$  in various spectral layers in FNO to achieve optimal generalization performance. Too few number of modes leads to underfitting, while too many can result in overfitting. Further, we need to train on high-enough resolution data to capture all the physical dynamics and approximate the ground-truth solution operator faithfully.

{\bf Our approach: }We propose the Incremental Fourier neural operator (iFNO) to address the above challenges.
iFNO incrementally and automatically increases the number of frequency modes $K$ and resolution $R$ during training.
It leverages the explained ratio as a metric to dynamically increase the number of frequency modes.
The explained ratio characterizes the amount of information in the underlying spectrum that the current set of modes can explain.
A small explained ratio (i.e., below a fixed threshold) indicates that the current modes are insufficient to represent the target spectrum and that more high-frequency modes should be added, as illustrated in Figure~\ref{fig:ifno_framework}. Simultaneously, iFNO also starts by first training on low-resolution data and progressively increases the resolution,  akin to the principle of curriculum learning, thereby increasing training efficiency.

\textbf{Our contributions are summarized as follows:}
    \vspace{-3mm}
\begin{enumerate}[leftmargin=*]
    \item We introduce the Incremental Fourier Neural Operator (iFNO), a novel approach that progressively increases the number of spectral coefficients and the data resolution used during training. These can be applied jointly or individually.
    \vspace{-2mm}
    \item We perform a thorough empirical validation across a range of partial differential equation (PDE) problems, including Burgers, Darcy Flow, the Navier-Stokes Equation, and the Kolmogorov Flow.
    \vspace{-2mm}
    \item Using our proposed iFNO, we demonstrate a 10\% lower testing error by acting as a dynamical spectral regularization scheme, using 20\% fewer frequency modes compared to the existing Fourier Neural Operator, while also achieving a 30\% faster performance, enabling larger scale simulations.
\end{enumerate}

%% file: figures/wrap_ifno_framework.tex
\begin{figure*}[t!]
    \centering
    \includegraphics[width=\textwidth]{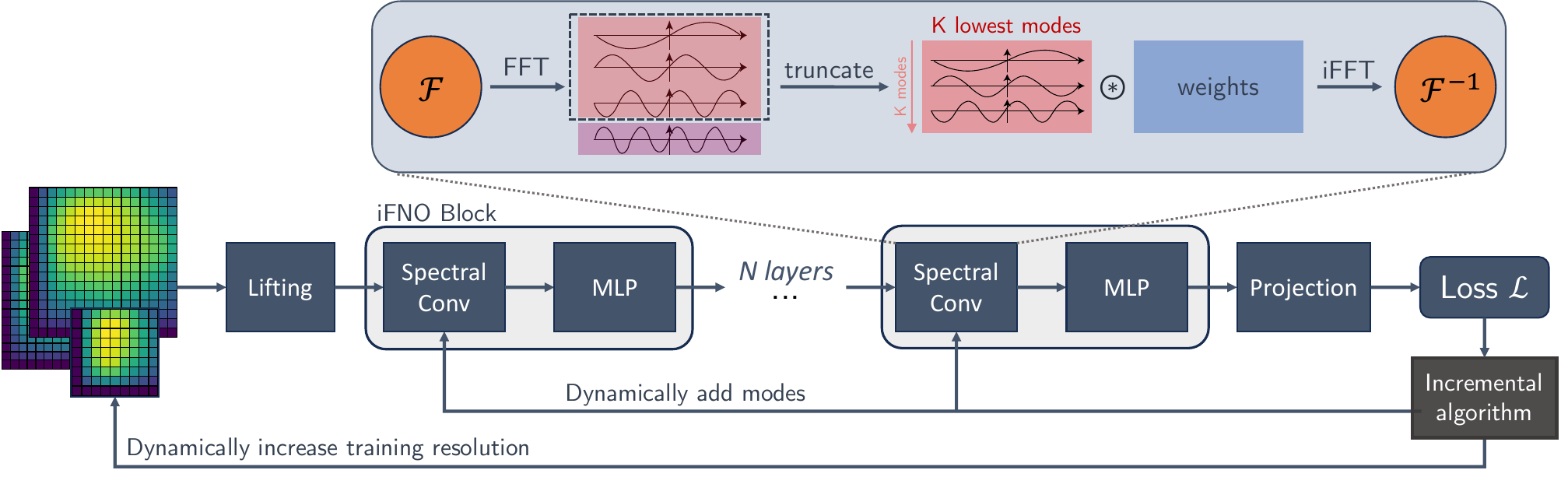}
    \caption{\small{
    \textbf{Top: Fourier convolution operator in FNO.} After the Fourier transform $\mathcal{F}$, the layer first truncates the full set of frequencies to the $K$ lowest ones using a dynamically set truncation before applying a learnable linear transformation (\emph{blue}) and finally mapping the frequencies back to the linear space using the inverse FFT $\mathcal{F}^{-1}$. The previous method \citep{li_fourier_2021} picks a fixed $K$ throughout the entire training. \textbf{Bottom: full iFNO architecture.} Our model takes as input functions at different resolutions (discretizations). The operator consists of a lifting layer, followed by a series of iFNO blocks. The loss is used by our method to dynamically update the input resolution and the number of modes $K$ in the Spectral Convolutions. The incremental algorithm is detailed in section~\ref{sec:ifno} and algorithm~\ref{alg:combined}.
    }}
    \label{fig:ifno_framework}
\end{figure*}

%% file: sections/related_works.tex
\section{Related Works}
The applications of neural networks on partial differential equations have a rich history~\citep{lagaris1998artificial,dissanayake1994neural}. 
These deep learning methods can be classified into three categories: (1) ML-enhanced numerical solvers such as learned finite element, finite difference, and multigrid solvers \citep{kochkov2021machine,pathak2021ml, greenfeld2019learning} ; (2) Network network-based solvers such as Physics-Informed Neural Networks (PINNs), Deep Galerkin Method, and Deep Ritz Method \citep{raissi2019physics,sirignano2018dgm,weinan2018deep}; and (3) the data-driven surrogate models such as \citep{guo2016convolutional,zhu2018bayesian,bhatnagar2019prediction}. 
Among them, the machine learning surrogate models directly parameterized the target mapping based on the dataset. They do not require a priori knowledge of the governing system and usually enjoy a light-fast inference speed. Recently, a novel class of surrogate models called neural operators was developed \citep{li_neural_2020,lu2021learning,kissas2022learning}. Neural operators parameterize the PDEs' solution operator in the function spaces and leverage its mathematical structures in their architectures. Consequentially, neural operators usually have a better empirical performance \citep{takamoto2022pdebench} and theoretical guarantees~\cite{kovachki2021neural} combined with conventional deep learning models.

The concept of implicit spectral bias was first proposed by \citet{rahaman_spectral_2019} as a possible explanation for the generalization capabilities of deep neural networks. 
There have been various results towards theoretically understanding this phenomenon \cite{cao_towards_2020,basri_convergence_2019}.
\citet{fridovich-keil_spectral_2022} further propose methodologies to measure the implicit spectral bias on practical image classification tasks. 
Despite a wealth of theoretical results and empirical observations, there has been no prior research connecting the implicit spectral bias with FNO to explain its good generalization across different resolutions.

The notion of incremental learning has previously been applied to the training of PINNs~\cite{characterize_failure_modes,huang_pinnup_2021}. 
However, these focus on a single PDE instance and apply incremental learning to the complexity of the underlying PDE, e.g., starting with low-frequency wavefields first and using high-frequency ones later~\cite{huang_pinnup_2021}. Our work is orthogonal to this direction, and rather than modifying the underlying PDE, we directly incrementally increase the model's capacity and the data's resolution.

%% file: sections/FNO/model_intro.tex
\section{Fourier Neural Operator}

Fourier Neural Operators belong to the family of neural operators, which are formulated as a generalization of standard deep neural networks to operator setting \citep{li_neural_2020}.
A neural operator learns a mapping between two infinite dimensional spaces from a finite collection of observed input-output pairs.
Let $D$ be a bounded, open set and $\mathcal{A}$ and $\mathcal{U}$ be separable Banach spaces of functions of inputs and outputs.

We want to learn a neural operator  $\mathcal{G}_\theta: \mathcal{A} \times \theta \to \mathcal{U}$ that maps any initial condition $a \in \mathcal{A}$ to its solution $u \in \mathcal{U}$.
The neural operator $\mathcal{G}_\theta$ composes linear integral operator $\mathcal{K}$ with pointwise non-linear activation function $\sigma$ to approximate highly non-linear operators. 

\begin{definition}[Neural operator]
    The neural operator $\mathcal{G}_\theta$ is defined as follows:
    $$
    \mathcal{G}_\theta:=\mathcal{Q} \circ\left(W_L+\mathcal{K}_L\right) \circ \cdots \circ \sigma\left(W_1+\mathcal{K}_1\right) \circ \mathcal{P},
    $$
    where $\mathcal{P}, \mathcal{Q}$ are the pointwise neural networks that encode the lower dimension function into higher dimensional space and decode the higher dimension function back to the lower dimensional space. The model stack $L$ layers of $\sigma\left(W_l+\mathcal{K}_l\right)$ where $W_l$ are pointwise linear operators (matrices), $\mathcal{K}_l$ are integral kernel operators, and $\sigma$ are fixed activation functions. The parameters $\theta$ consist of all the parameters in $\mathcal{P}, \mathcal{Q}, W_l, \mathcal{K}_l$.
\end{definition}

\citet{li_fourier_2021} proposes FNO that adopts a convolution operator for $\mathcal{K}$ as shown in Figure~\ref{fig:ifno_framework}, which obtains state-of-the-art results for solving PDE problems.
\begin{definition}[Fourier convolution operator]
    Define the Fourier convolution operator $\mathcal{K}$ as follows:
    $$
    \left(\mathcal{K} v_t\right)(x)=\mathcal{F}^{-1}\left(R \cdot T_K\left(\mathcal{F} v_t\right)\right)(x) \quad \forall x \in D,
    $$
    where $\mathcal{F}$ and $\mathcal{F}^{-1}$ are the Fourier transform and its inverse, $R$ is a learnable transformation and $T_K$ is a fixed truncation that restricts the input to lowest $K$ Fourier modes.
\end{definition}

FNO is discretization-convergent, such that the model can produce a high-quality solution for any query points, potentially not in the training grid. In other words, FNO can be trained on low-resolution but generalizes to high-resolution. This property is highly desirable as it allows a transfer of solutions between different grid resolutions and discretizations.

\paragraph{Frequency truncation.}
To ensure discretization-convergence, FNO truncates the Fourier series $\mathcal{\hat{F}} v$ at a maximal number of modes $K$ using $T_K$.
In this case, for any discrete frequency mode $k \in \{1,...,K \}$,  we have $\mathcal{\hat{F}}v (k) \in \mathbb{C}^{\textrm{C}}$ and $R(k) \in \mathbb{C}^{\textrm{C} \times \textrm{C}}$, where $\textrm{C}$ is the channel dimension of the input $v$.
The size of linear parameterization $R$ depends on $K$.
For example, $R$ has the shape of $K^d \times \textrm{C} \times \textrm{C}$ for a $d$-dimensional problem. 
We also denote the modes $1,..., K$ as the effective frequency modes. 
In the standard FNO, \citet{li_fourier_2021} view $K$ as an additional hyperparameter to be tuned for each problem. 

\paragraph{Frequency strength.}
As $R(k)$ reflects how the frequency mode $k$ is transformed, we can measure the strength of transforming the frequency mode $k$ by the square of the Frobenius norm of $R(k)$, such that:
\begin{equation}
    \label{eq:freq_strength}
    S_{k} = \sum_{i}^{\textrm{C}} \sum_{j}^{\textrm{C}} |R_{k, i, j}|^2,
\end{equation}
where $S_{k}$ denotes the strength of the $k$-th frequency mode.
A smaller $S_{k}$ indicates that the $k$-th frequency mode is less important for the output.

\begin{figure}[t!]
    \centering
    \hspace*{-0.8cm}
    \includegraphics[width=0.9\columnwidth]{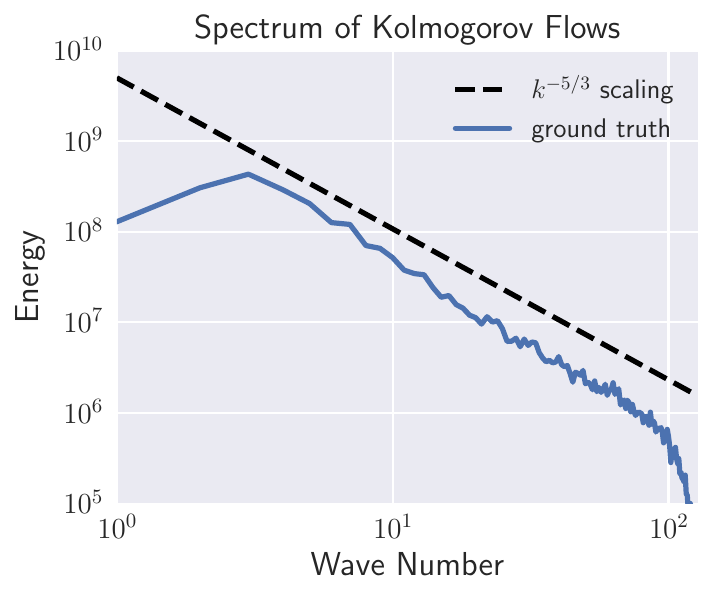}
    \vskip -0.3cm
    \caption{\small{The spectrum of Kolmogorov flow decays exponentially with the Kolmogorov scale of $5/3$ in the inverse cascade range.}}
    \label{fig:spectrum}
    \vskip -0.5cm
\end{figure}

\begin{figure*}[t!]
    \centering
    \includegraphics[width=0.8\textwidth]{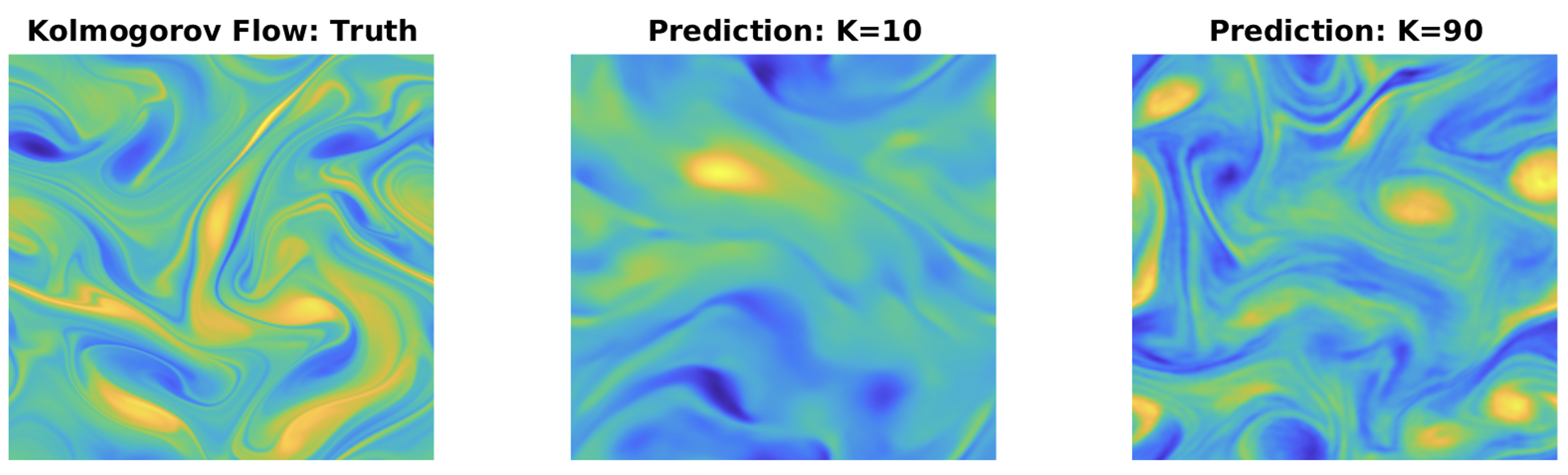}
    \caption{\small{FNO with higher frequency modes captures smaller-scale structures in the turbulence.
    Prediction of the Kolmogorov flow by FNO with $K=10$ and $K=90$.
    Insufficient modes lead to overly strong dumping and it fails to capture small-scale structures in the turbulence of Kolmogorov flow.
    }}
    \label{fig:re5000}
\end{figure*}

%% file: sections/importance.tex
\paragraph{Implicit spectral bias in neural networks}
It has been well studied that neural networks implicitly learn low-frequency components first, and then learn high-frequency components in a later stage \cite{rahaman_spectral_2019,xu2019frequency}.
This phenomenon is known as implicit spectral bias, and it helps explain the excellent generalization ability of overparameterized neural networks.

Since FNO performs linear transformation $R$ in the frequency domain, the frequency strength $S_k$ of each frequency mode $i$ is directly related to the spectrum of the resulting model.
As FNO is a neural network and trained by first-order learning algorithms, it follows the implicit spectral bias such that the lower frequency modes have larger strength in $R$. 
This explains why FNO chooses to preserve a set containing the lowest frequency modes, instead of any arbitrary subset of frequencies.

\input{sections/ifno_algo}

\paragraph{Low-frequency components in PDEs}

Learning frequency modes is important as large-scale, low-frequency components usually have larger magnitudes than small-scale, high-frequency components in PDEs.
For dissipative systems (with diffusion terms) such as the viscous Burgers' equation and incompressible Navier-Stokes equation, the energy cascade involves the transfer of energy from large scales of motion to the small scales, which leads to the Kolmogorov spectrum with the slope of $k^{-5/3}$ in the inverse cascade range (Figure \ref{fig:spectrum}), and $k^{-3}$ in the direct-cascade range \citep{boffetta2012two}. The smallest scales in turbulent flow is called the Kolmogorov microscales. 
Therefore, one should choose the model frequencies with respect to the underlying equation frequencies when designing machine learning models.
It would be a challenge to select the correct model frequencies in advance, without knowing the properties of the underlying PDEs.

%% file: sections/ifno_algo.tex
\begin{algorithm*}[ht]
\caption{\textit{Incremental Learning of FNO}}
\label{alg:combined}
\begin{algorithmic}
\STATE {\bfseries Input:} Initial modes $K_0$, mode buffers $b$, threshold $\alpha$, epoch $e$ to increase resolution, downsample factors $r$, data resolution $R_{data}$
\STATE {\bfseries Initialize:} Randomly initialize $R_0 \in \Complex^{(K_0+b) \times C \times C}$, set initial resolution $R_{curr} = R_{data}/r_1$
\STATE{\bfseries For iteration $t$ in $1,...,T$:}
\STATE{\bfseries \quad} compute $s_t$ = $[S_1, S_2, ..., S_{K_{t-1}+b}]$ \COMMENT{compute the frequency strength for each mode}
\STATE{\bfseries \quad} $K_t \gets K_{t-1}$
\STATE{\bfseries \quad While $g(K_t, s_t) < \alpha$:} \COMMENT{find $K_t$ that explains at least $\alpha$ of $s_t$}
\STATE{\bfseries \quad \quad} $K_t \gets K_t + 1$
\STATE{\bfseries \quad} construct $R_t \in \mathbb{C}^{(K_t+b) \times\textrm{C} \times\textrm{C}}$ where $R_t[0:(K_{t-1}+b),:,:] = R_{t-1}$ and initialize the rest randomly 
\STATE{\bfseries \quad If ($t \bmod e = 0$)}
\STATE{\quad \quad Set next resolution $R_{curr} = R_{data}/r_{i+1}$}
\STATE{\bfseries \quad Train FNO model with $(K_t, R_{curr})$}
\end{algorithmic}
\label{alg:combined}
\end{algorithm*}

%% file: sections/motivations.tex
\section{Incremental Fourier Neural Operator}\label{sec:ifno}

Although FNO has shown impressive performance in solving PDEs, the training remains challenging. In this section, we discuss the main difficulties in training FNOs and propose the Incremental Fourier Neural Operator (iFNO) to address the challenges.

\subsection{Incremental frequency FNO}

While frequency truncation $T_K$ ensures the discretization-invariance property of FNO, it is still challenging to select an appropriate number of effective modes $K$, as it is task-dependent and requires careful hyperparameter tuning.

Inappropriate selection of $K$ can lead to severe performance degradation.
Setting $K$ too small will result in too few frequency modes, such that it doesn't approximate the solution operator well, resulting in underfitting.
As shown in Figure~\ref{fig:re5000}, in the prediction of Kolmogorov flow, FNO requires higher frequency modes to capture small-scale structures in turbulence for better performance.
On the other hand, a larger $K$ with too many effective frequency modes may encourage FNO to interpolate the noise in the high-frequency components, leading to overfitting.

Previously, people have chosen the number of effective modes in FNO by estimating the underlying data frequencies, or using heuristics borrowed from pseudo-spectral solvers such as the 2/3 dealiaising rule \citep{hou2007computing}, i.e., picking the max Fourier modes as 2/3 of the training data resolution. 
However, it's usually not easy to estimate the underlying frequency, and the resolution of data may vary during training and testing. These problems lead to hand-tuning or grid-searching the optimal effective modes, which can be expensive and wasteful. We propose the \textit{incremental Fourier Neural Operator (iFNO)} that starts with a "small" FNO model with limited low-frequency modes and gradually increases $K$ and resolution $R$ based on the training progress or frequency evolution. One of the key challenges in training iFNO is determining an appropriate time to proceed with the increase of $K$. We consider two variants of iFNO (method), which means we only increase the frequency modes based on the method.

\vspace{-3mm}
\paragraph{iFNO (Loss):}A common practice is to use the training progress as a proxy to determine the time, as seen in previous work such as \citet{liu_splitting_2019}.
Specifically, we let the algorithm increase $K$ only decrease in training loss between $N$ consecutive epochs is lower than a threshold $\epsilon$.
We denote it as the iFNO (loss) algorithm.
Despite its simplicity, this heuristic has shown to be effective in practice. 
However, finding an appropriate $\epsilon$ and $N$ is challenging as it depends on the specific problem and the loss function.
To address this issue, we propose a novel method to determine the expanding time by directly considering the frequency evolution in the parameter space.
\vspace{-3mm}
\input{figures/wrap_freq_evolution.tex}

\paragraph{iFNO (Freq):}As shown in Equation~\ref{eq:freq_strength}, $i$-th frequency strength $S_i$ reflects the importance of the $i$-th frequency mode in the model.
The entire spectrum of the transformation $R$ can be represented by the collection of $p$ frequency strengths (when $p$ is sufficiently large):
\begin{equation}
    \label{eq:freq_strength_dist}
    \begin{aligned}
        s_p = [S_1,S_2,..., S_p],
    \end{aligned}
\end{equation}
where $s_p \in \mathbb{R}^{p}$. 
When applying the frequency truncation $T_K$ ($K < p$), we can measure how much the lowest $K$ frequency modes explain the total spectrum by computing the \emph{explanation ratio} $g(K,s_p)$.
\begin{equation}
    \label{eq:explanation_ratio}
    g(K,s_p) =  \frac{\sum_{k=1}^{K} S_k}{\sum_{k=1}^{p} S_k}.
\end{equation}    
We define a threshold $\alpha$, which is used to determine if the current modes can well explain the underlying spectrum.
If $g(K,s_p) > \alpha$, it indicates that the current modes are sufficient to capture the important information in the spectrum, and thus no additional modes need to be included in FNO.
Otherwise, more frequency modes will be added into the model until $g(K,s_p) > \alpha$ is satisfied.

Although $s_p$ reflects the entire spectrum when $p$ is sufficiently large, maintaining a large $p$ is unnecessary and computationally expensive.
Instead, we only maintain a truncated spectrum $s_{K+b} \in \mathbb{R}^{K+b}$, which consists of $K$ effective modes and $b$ buffer modes.
The buffer modes contain all mode candidacies that potentially would be included as effective modes in the later stage of training.

In practice, at iteration $t$ with $K_t$ effective modes, we construct a transformation $R_t \in \mathbb{C}^{(K_t+b) \times\textrm{C} \times\textrm{C}}$ for 1D problems. 
After updating $R_t$ at iteration $t+1$, we will find $K_{t+1}$ that satisfies $g(K_{t+1},s_t) > \alpha$.
In this way, we can gradually include more high-frequency modes when their evolution becomes more significant, as illustrated in Figure \ref{fig:ifno_framework}. We denote this variant as iFNO (Freq). We also describe how iFNO (Freq) is trained for solving 1D problems in Algorithm \ref{alg:combined}. iFNO is extremely efficient as the training only requires a part of frequency modes.
In practice, it also converges to a model requiring fewer modes than the baseline FNO without losing any performance, making the model efficient for inference.
Notably, the cost of computing the frequency strength and explanation ratio is negligible compared to the total training cost. 
It can be further reduced by performing the determination every $T$ iterations, which does not affect the performance. 

\begin{table*}[ht]
    \centering
    \begin{threeparttable}
        \caption{\textbf{Evaluation on Re5000 dataset on full data regime}. The average training L2 loss and testing L2 loss on super-resolution are reported on 3 runs with their standard deviation.}
        \label{tab:re5000}
        \begin{tabular}{lccccc}
            \toprule
            Method & Train L2 & Test L2 (Super-Res) (1e-1) & Training Time (Mins) \\
            \midrule
            iFNO &  $\textbf{0.149} \pm \textbf{0.004}$ & $0.961 \pm 0.022$ & \textbf{670} $\pm$ \textbf{40} \\
            Standard FNO & 0.169 $\pm$ 0.004 & $0.988 \pm 0.031$ & $918 \pm 10$ \\
            iFNO (Freq) & $0.163 \pm 0.003$ & 1.019 $\pm$ 0.031 & $700 \pm 30$ \\
            iFNO (Loss) & $0.155 \pm 0.005$ 
            & $\textbf{0.948} \pm \textbf{0.040}$ & $690 \pm 30$ \\
            iFNO (Resolution only) & $0.155 \pm 0.003$ & $0.974 \pm 0.010$ & $668 \pm 40$\\
            \bottomrule
        \end{tabular}
    \end{threeparttable}
\end{table*}

\vspace{-3mm}

\subsection{Regularization in FNO training}
\label{sec:fno_regularization}

As discussed in the previous section, learning low-frequency modes is essential for the successful training of FNO. 
However, the standard regularization techniques used in training neural networks are not capable of explicitly promoting the learning of low-frequency modes.

As shown in Figure \ref{fig:frequency_evolution},  although FNO with well-tuned weight decay strength successfully regularizes the high-frequency modes, its regularization can be too strong for certain layers (such as the fourth operator in the figure).
This leads to the instability of the frequency evolution, which damages the generalization performance, as shown in Table~\ref{tab:main_results}.
On the other hand, insufficient decay strength cannot regularize the high-frequency modes and can even lead to overfitting the associated high-frequency noise. 
\vspace{-3mm}
\paragraph{Dynamic Spectral Regularization:} However, we find that iFNO can be served as a dynamic spectral regularization process.
As shown in Figure~\ref{fig:frequency_evolution}, iFNO properly regularizes the frequency evolution without causing instability or overfitting high-frequency modes.
As we will present in the experiment section, the dynamic spectral regularization gives iFNO a significant generalization improvement.

\subsection{Incremental resolution FNO}

Finally, we introduce the incremental algorithm on the resolution part of our approach.

\paragraph{Computational cost in training large-scale FNO}
Training the FNO on large-scale, high-resolution data to solve PDEs with high-frequency information, a large effective mode number $K$ is necessary to capture the solution operator. However, this requires constructing a large-scale transformation $R$, which dominates the computational cost of all the operations in FNO. 
This makes the training and inference more computationally expensive.
For example, the Forecastnet \citep{pathak2022fourcastnet} for global weather forecast based on FNO is trained on 64 NVIDIA$^{\circledR}$ Tesla$^{\circledR}$  A100 GPUs, only covering a few key variables. To simulate all the weather variables, it potentially needs the parallel FNO, which scales to 768 GPUs~\citep{grady2022towards}.

\paragraph{Incremental Resolution:} To mitigate the computational cost of training FNO on high-resolution data, we incorporate curriculum learning to improve our model (\citet{soviany2022curriculum}). This approach involves training the models in a logical sequence, where easier samples are trained first and gradually progressing to more challenging ones. Our samples are images; hence, we can easily distinguish between easy and difficult samples based on their resolution. Higher-resolution images are more complex to train on, so it's best to start with lower-resolution samples and gradually move up to higher-resolution ones.  We can achieve better performance than standard training methods while reducing the computational burden of training the entire model. This approach allows the model to initially focus on learning low-frequency components of the solution using small data, avoiding wasted computation and overfitting. Then, the model progressively adapts to capture higher frequency components as the data resolution increases. This model combines both incremental frequency (any of the two models) and incremental resolution and is denoted by iFNO. Algorithm \ref{alg:combined} showcases the joint implementation.

%% file: figures/wrap_freq_evolution.tex
\begin{figure*}[t!]
    \centering
    \includegraphics[width=1.0\linewidth]{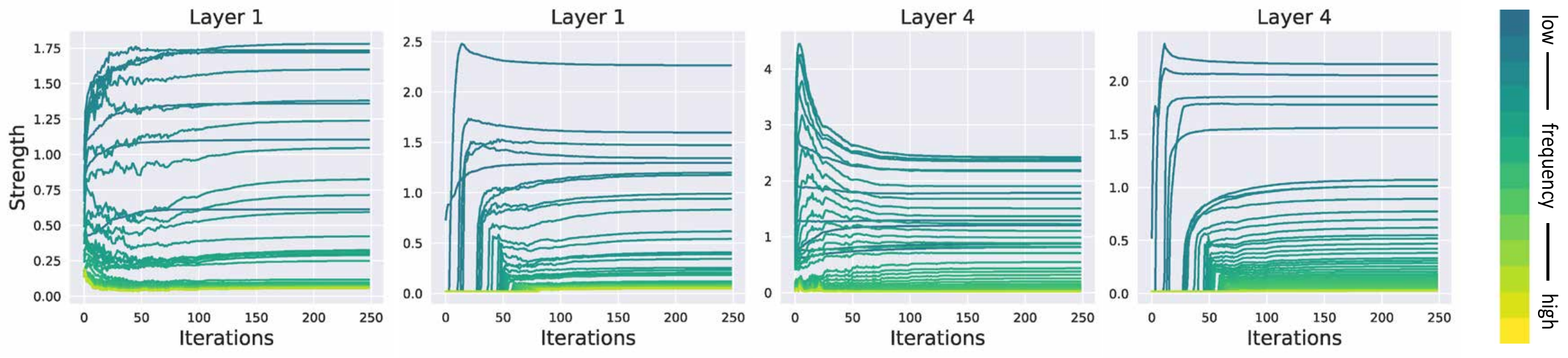}
    \caption{\small{Frequency evolution of first and fourth Fourier convolution operators in FNO and iFNO during the training on Burgers' equation. We visualize FNO on the left figure and iFNO on the right figure for each layer. Each frequency strength $S_k$ is visualized across training. FNO is tuned with the optimal weight decay strength.
    }}
    \vskip -0.1in
    \label{fig:frequency_evolution}
\end{figure*}

%% file: sections/experiments.tex
\section{Evaluations}

In this section, we detail the experimental setting, datasets used and analyze the empirical results.

\begin{table*}[t!]
\centering
  \caption{\small{\textbf{Evaluation on various PDEs in the low-data regime}. The average training and testing L2 loss are reported with their standard deviation over 3 repeated runs. Frequency-based and loss-based iFNO (frequency modes only) are compared with standard FNO with fixed $K = 10, 30, 60, 90$ number of modes. Losses have been divided by the scale under the column. 
  }}
\begin{tabular}{lccccccc}
    \toprule
    \textbf{Method} &
      \multicolumn{2}{c}{\textbf{Burgers}} &
      \multicolumn{2}{c}{\textbf{Darcy Flow}}  &
      \multicolumn{2}{c}{\textbf{Navier Stokes (FNO2D)}} \\
      & {Train (\num{1e-3})} & {Test (\num{1e-1})} & {Train (\num{1e-3})} & {Test (1e-5)} & {Train (\num{1e-3})} & {Test (1)}\\
      \cmidrule(r){2-3}\cmidrule(r){4-5}\cmidrule(r){6-7}
        iFNO (Freq) & $0.310 \pm 0.050$ & \textbf{0.874} $\pm$ \textbf{0.044} & \textbf{0.059} $\pm$ \textbf{0.001} & \textbf{1.127} $\pm$ \textbf{0.034} & $142.3 \pm 6.621$ & \textbf{0.589} $\pm$ \textbf{0.017}\\
        iFNO (Loss) & $13.53 \pm 13.08$ & $1.197 \pm 0.398$ & {0.059} $\pm$ {0.001} & $1.128 \pm 0.025$ & $126.8 \pm 17.51$ & $0.616 \pm 0.017$\\
        \cmidrule(r){2-7}
        FNO (10) & $2.844 \pm 0.070$ & $0.934 \pm 0.054$ & {0.061} $\pm$ {0.001} & $1.144 \pm 0.020$ & $139.9 \pm 6.913$ & $0.618 \pm 0.004$\\
        FNO (30) & $0.276 \pm 0.083$ & $0.960 \pm 0.004$ & {0.060} $\pm$ {0.003} & $1.123 \pm 0.011$ & $118.6 \pm 11.58$ & $0.638 \pm 0.018$\\
        FNO (60) & $0.316 \pm 0.015$ & $0.899 \pm 0.028$ & {0.060} $\pm$ {0.002} & $1.138 \pm 0.005$ & \textbf{87.64} $\pm$ \textbf{8.016} & $0.644 \pm 0.013$\\
        FNO (90) & \textbf{0.158} $\pm$ \textbf{0.064} & $0.896 \pm 0.022$ & {0.059} $\pm$ {0.001} & $1.130 \pm 0.023$ & $113.7 \pm 12.12$ & $0.645 \pm 0.022$\\
    \bottomrule
  \end{tabular}
  \vskip 0.1in
\centering
\begin{tabular}{lccccc}
    \toprule
    \textbf{Method} &
      \multicolumn{2}{c}{\textbf{Navier Stokes (FNO3D)}} &
     \multicolumn{2}{c}{\textbf{Kolmogorov Flow}}\\
      & {Train (\num{1e-3})} & {Test (\num{1e-1})} & {Train (\num{1e-3})} & {Test (\num{1e-1})}\\
      \cmidrule(r){2-3}\cmidrule(r){4-5}
        iFNO (Freq) & $4.350 \pm 0.194$ & $5.575 \pm 0.039$ & {36.25} $\pm$ {0.533} & $1.290 \pm 0.022$\\
        iFNO (Loss) & $2.083 \pm 0.277$ & $5.605 \pm 0.051$ & {34.44} $\pm$ {1.075} & \textbf{1.275} $\pm$ \textbf{0.015}\\
        \cmidrule(r){2-5}
        FNO (10) & $5.393 \pm  0.070$ & \textbf{5.570} $\pm$ \textbf{0.014} & {142.3} $\pm$ {0.432} & $1.792 \pm 0.003$ \\
        FNO (30) & \textbf{1.895} $\pm$  \textbf{0.273} & $5.627 \pm 0.073$ & {69.91} $\pm$ {0.511} & $1.392 \pm 0.022$\\
        FNO (60) & $-$ & $-$ & {38.59} $\pm$ {0.333} & $1.386 \pm 0.023$\\
        FNO (90) & $-$ & $-$ & \textbf{29.05} $\pm$ \textbf{1.066} & $1.387 \pm 0.015$\\
    \bottomrule
  \end{tabular}
  \label{tab:main_results}
\end{table*}

\subsection{Experimental Setup}

We evaluate iFNO and its variants on five different datasets of increasing difficulty. We retain the structure of the original FNO, which consists of stacking four Fourier convolution operator layers with the ReLU activation and batch normalization. More specifically, for the time-dependent problems, we use an RNN structure in time that directly learns in space-time\footnote{We share our code via \href{https://anonymous.4open.science/r/icml-ifno/}{an anonymous link here}.}.
The initial and coefficient conditions are sampled from Gaussian random fields \citep{nelsen2021random} for all the datasets.
\vspace{-4mm}
\paragraph{Kolmogorov Flow:} We evaluate the models on a very challenging dataset of high-frequency Kolmogorov Flow. The Re5000 problem is the most challenging dataset we consider.  It is governed by the 2D Navier-Stokes equation, with forcing 
$f(x) = \mathrm{sin}(nx_2) \hat{x}_1$,
where $x_1$ is the unit vector and a larger domain size $[0,2\pi]$, as studied in \citep{li2021markov, kochkov2021machine}.
More specifically, the Kolmogorov flow (a form of the Navier-Stokes equations) for a viscous, incompressible fluid,
\begin{align*}
\frac{\partial u}{\partial t}=-u \cdot \nabla u-\nabla p+\frac{1}{R e} \Delta u+\sin (n y) \hat{x}, \quad \nabla \cdot u=0, \\ 
\quad \text { on }[0,2 \pi]^2 \times(0, \infty)
\end{align*}
with initial condition $u(\cdot, 0)=u_0$ where $u$ denotes the velocity, $p$ the pressure, and $R e>0$ is the Reynolds number. It has an extremely high Reynolds number with $Re = 5000$. The resolution of this dataset is $128 \times 128$. We consider learning the evolution operator from the previous time step to the next based on 2D FNO. We also consider a variation of the dataset with a lower $Re$ value, which we call the Kolmogorov Flow problem. This resolution is fixed to be $256 \times 256$ to show high-frequency details. 

\textbf{Burgers' Equation}: A dataset of a non-linear PDE with various applications, including modeling the flow of a viscous fluid. 

\textbf{Darcy Flow}: The steady-state of the 2D Darcy Flow on the unit box, which is a second-order, linear, elliptic PDE. 

\textbf{2D and 3D Navier-Stokes Equation}: We consider solving a 2D+time Navier-Stokes equation using 2D FNO with an RNN structure in time or using the 3D FNO to do space-time convolution. The above dataset settings are similar to the settings in \citet{li_fourier_2021}, which are listed in the appendix.

\paragraph{Methods}
We present our results for each dataset where we consider four standard FNO baselines with different numbers of frequency modes $K=10,30,60,90$.
For each dataset, all methods are trained with the same hyperparameters, including the learning rate, weight decay, and the number of epochs.
Unless otherwise specified, we use the Adam optimizer to train for 500 epochs with an initial learning rate of 0.001, which is halved every 100 epochs, and weight decay is set to be 0.0005. 
For the re5000 dataset, we train for 50 epochs, and experiment details are explained in the Appendix. All experiments are run using NVIDIA$^{\circledR}$ Tesla$^{\circledR}$ V100/A100 GPUs, and we repeat each experiment with 3 different random seeds. We also test our incremental methods on Tensorized Factorized Neural Operators~\cite{kossaifi2023multi} by sweeping over different factorization schemes and ranks for the weights.

\paragraph{Hyperparameter settings}
We find iFNO is \emph{less sensitive} to the choice of hyperparameters across different datasets. This property allows the same iFNO architecture and hyperparameters to be directly applied to a wide range of problems without expensive parameter validation and without needing to set an appropriate $K$.
Consequently, we use the same hyperparameters for all datasets, including the number of initial modes $K_0=1$, the number of buffer modes $b=5$, and the threshold $\alpha=0.99$.
We also use a uniform threshold $\epsilon=0.001$ to determine whether the training progress is stalled in iFNO (Loss). 

\subsection{Results}

\input{figures/wrap_main_results.tex}
First, we study the performance of our model on the large-scale Re5000 dataset, where we use 36000 training samples and 9000 testing samples. 
The results in Table \ref{tab:re5000} show that iFNO achieves significantly better generalization performance than just incrementally increasing either frequency or resolution alone. This highlights the advantages of the iFNO while gaining around a 30\% efficiency and much better accuracy than the stand-alone Incremental methods. The training time is reduced due to starting with a very small model and input data. There is one important note to mention that, in fact, for the Re5000 dataset, since the problem is highly challenging, i.e., having a high Reynolds number, we need to reach all 128 modes to capture even the high-frequency components. 

Next, we evaluate all methods in the low-data regime across different datasets, where the models only have access to a few data samples.
This is common in many real-world PDE problems where accurately labeled data is expensive to obtain.
In this regime, the ability to generalize well is highly desirable.
We consider a small training set (5 to 50 samples) for Burgers, Darcy, and Navier-Stokes datasets. As the data trajectories are already limited in the Kolmogorov dataset, we choose 8x larger time step (sampling rate) for each trajectory, resulting in fewer time frames. 

As shown in Table~\ref{tab:main_results}, iFNO (Freq) consistently outperforms the FNO baselines across all datasets, regardless of the number of modes $K$.
It also shows that FNO with larger $K$ achieves lower training loss but overfits the training data and performs poorly on the testing set.
On the other hand, iFNO (Freq) achieves slightly higher training loss but generalizes well to the testing set, which demonstrates the effectiveness of the dynamic spectral regularization.
We also notice that although iFNO (Loss) has significant fluctuations, especially in the training results, it does achieve better generalization on most datasets against the baselines. We also include the full data regime in the appendix, showing that on the Navier-Stokes and the Kolmogorov flow datasets, iFNO (Freq) achieves up to 6\% better generalization error compared to FNO.

\subsection{Ablation Studies}

\input{figures/wrap_loss_plots_epochs.tex}
\input{figures/wrap_bar_plot.tex}

Figure~\ref{fig:loss_plots_epochs} compares FNO (90) and iFNO (Freq) over the training on Kolmogorov flow. 
It shows that iFNO requires much fewer frequency modes during training.
As the linear transformation $R$ dominates the number of parameters and FLOPs in FNO when $K$ is large, iFNO (Freq) can significantly reduce the computational cost and memory usage of training FNO. iFNO achieves better generalization and even requires fewer modes at the end of training.
In Figure~\ref{fig:bar_plot}, we compare iFNO (Freq) with FNO with a particular $K$ that achieves the best performance in each dataset. 
The results suggest that the trained iFNO (Freq) consistently requires fewer frequency modes than its FNO counterpart, making the iFNO inference more efficient.
This also indicates that iFNO can automatically determine the optimal number of frequency modes $K$ during training without predetermining $K$ as an inductive bias that could potentially hurt the performance. Lastly, we show that TFNOs with our incremental approach (iTFNO) achieve up to 11\% better generalization error and up to 22\% performance speed up compared to baselines. The results are in the appendix.

%% file: figures/wrap_loss_plots_epochs.tex
\begin{figure}[t!]
    \centering
    \includegraphics[width=0.9\linewidth]{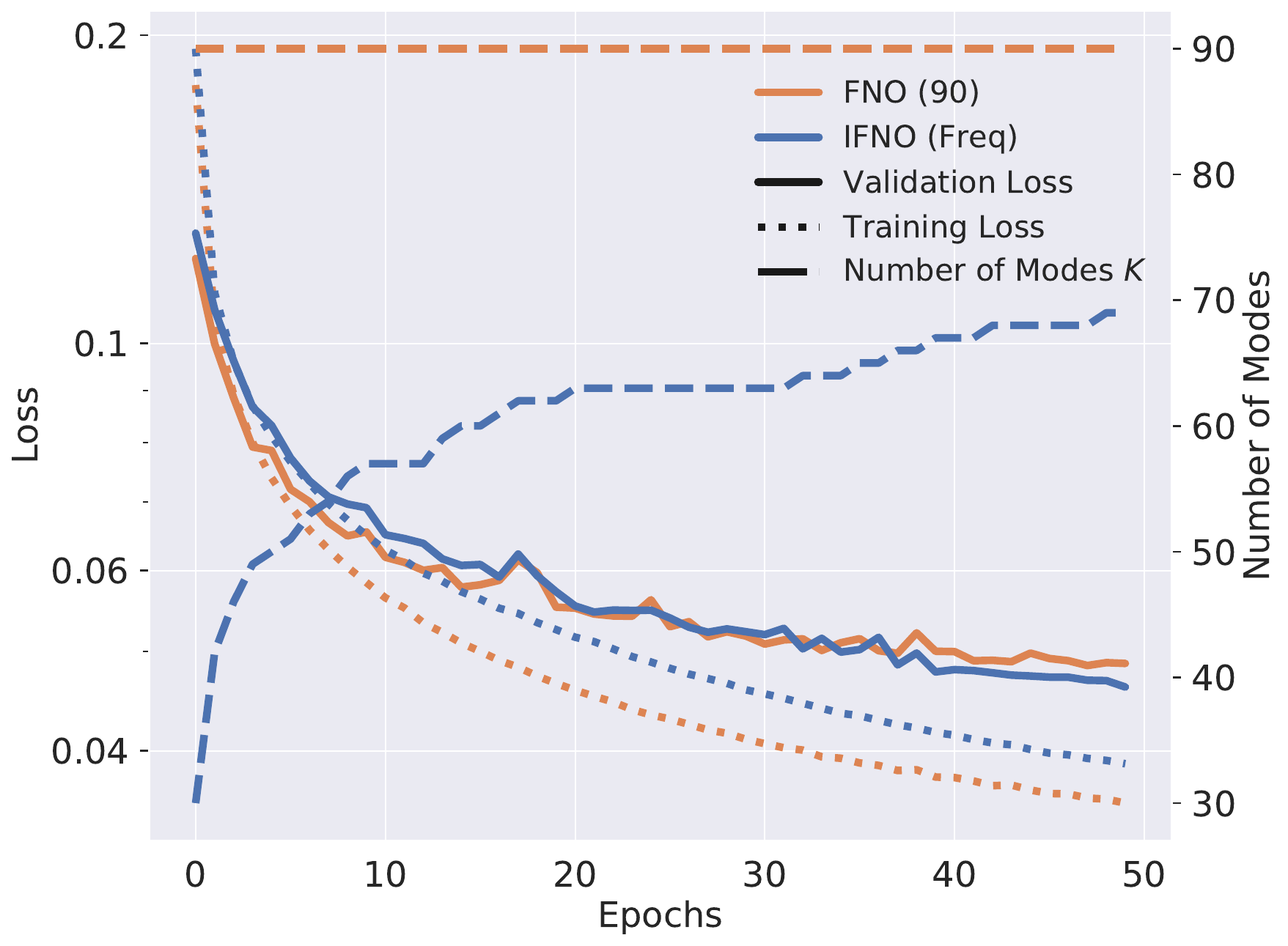}
    \vskip -0.1cm
    \caption{\small{Testing loss, training loss, and number of modes $K$ during the training of FNO and iFNO on Kolmogorov flow. 
    }}
    \label{fig:loss_plots_epochs}
    \vskip -0.1cm
\end{figure}

%% file: figures/wrap_bar_plot.tex
\begin{figure}[h!]
    \centering
    \includegraphics[width=0.80\linewidth]{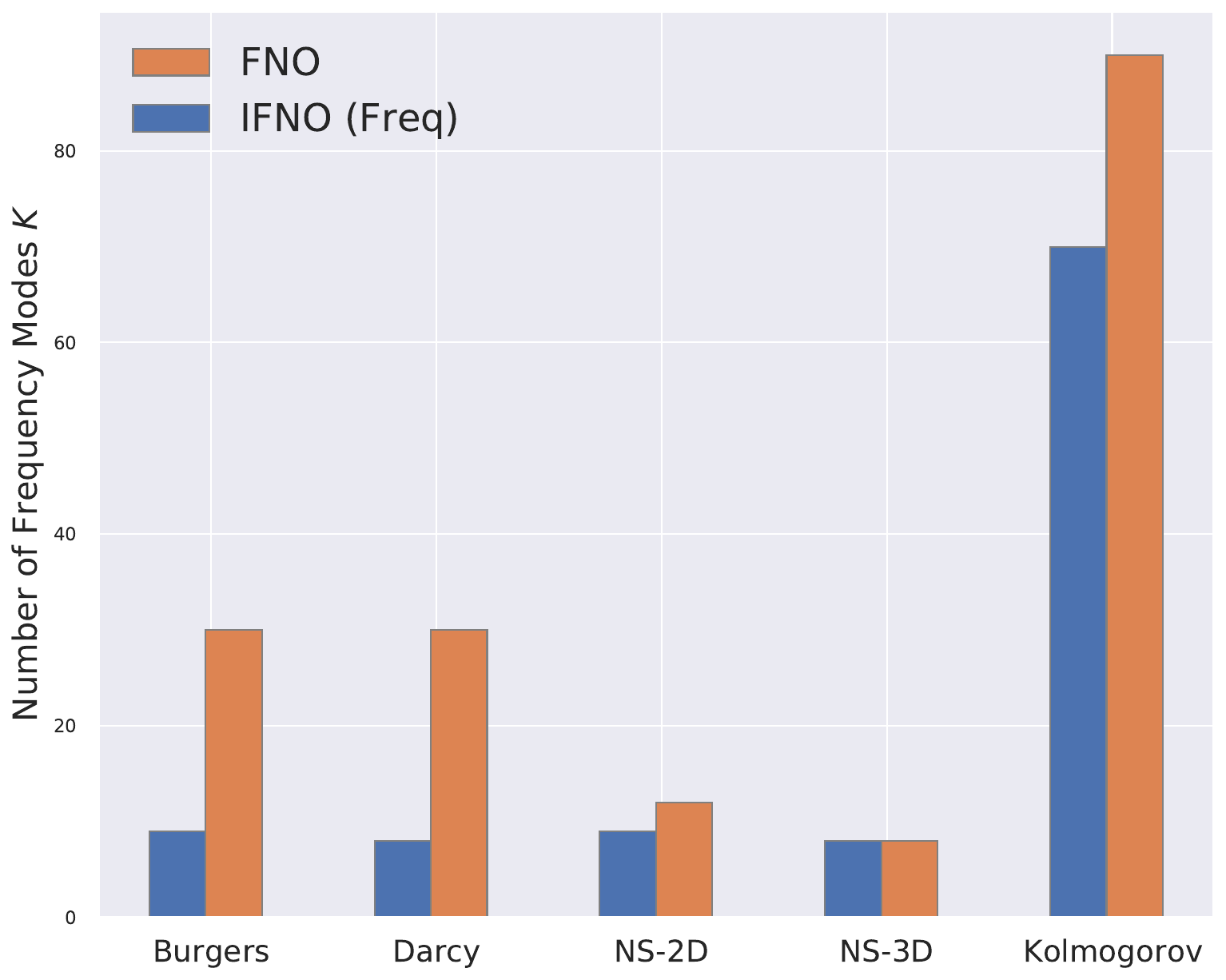}
    \caption{\small{Number of frequency modes $K$ in the converged FNO and iFNO models across datasets. We report $K$ in the first Fourier convolution operator. NS denotes Navier-Stokes equations.
    }}
    \label{fig:bar_plot}
    \vskip -0.4cm
\end{figure}

%% file: sections/conclusion.tex
\section{Conclusion}

In this work, we proposed the Incremental Fourier Neural Operator (iFNO) that dynamically selects frequency modes and increases the resolution during training. 
Our results show that iFNO achieves better generalization while requiring less number of parameters (frequency modes) compared to the standard FNO.
We believe iFNO opens many new possibilities, such as solving PDEs with very high resolution and with limited labeled data and training computationally efficient neural operators.
In the future, we plan to apply iFNO on more complex PDEs and extend it to the training of physics-informed neural operators.

\pagebreak

\section{Impact statement}
This paper presents work whose goal is to advance the field of Machine Learning and PDEs. FNOs have shown promise to solve challenging physics and engineering problems accurately and efficiently. Our research has made FNOs more optimized through incremental learning and, hence, could lead to FNOs being applied to solve larger real-world problems, allowing for advances in fields like engineering, weather forecasting, and climate science while reducing training times with fewer computing resources. There might also be potential negative societal consequences of our work, but none of them are immediate to be specifically highlighted here.

%% file: sections/appendix/discussions.tex
\section{Standard Regularization in training FNO}

\begin{figure*}[h!]
    \centering
    \includegraphics[width=0.9\linewidth]{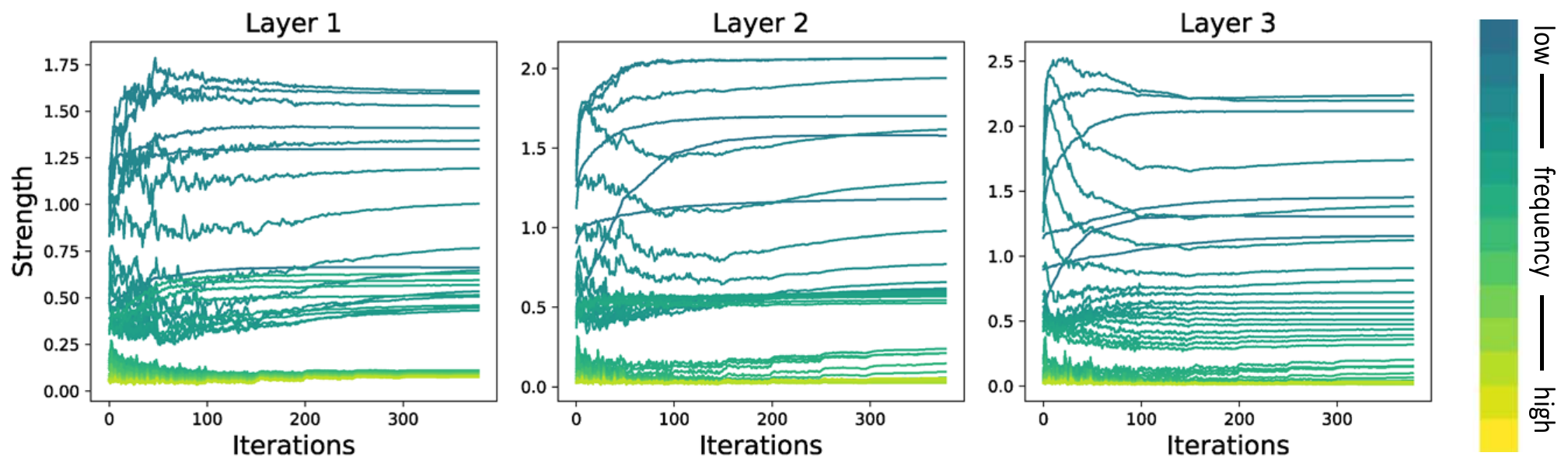}
    \includegraphics[width=0.9\linewidth]{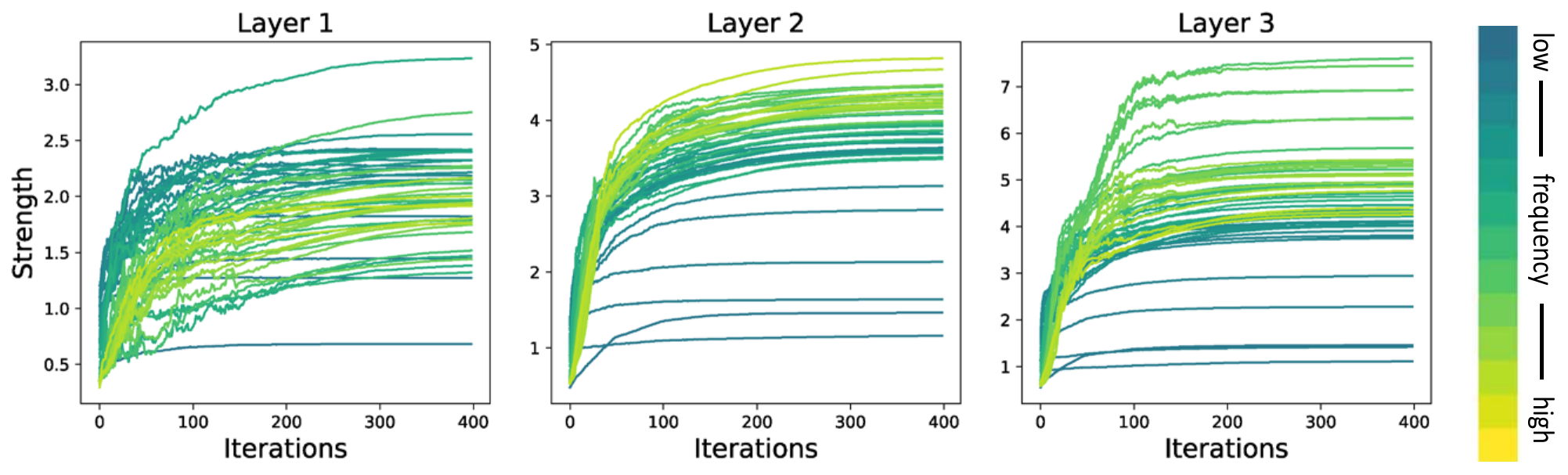}
    \caption{\small{
    Frequency evolution of the Fourier convolution operators in FNO with and without weight decay. \textbf{Top}: with weight decay; \textbf{bottom}: without weight decay. 
    When training with weight decay, the strength of each mode converges around 100 training iterations. And the weights of lower frequency tend to have high strength.
    On the other hand, the weight continues growing throughout the training without the weight decay. In this case, higher frequencies can also have high strength.
    }}
    \label{fig:weight-decay}
\end{figure*}

In this section, we visualize the frequency evolution of FNO and iFNO.
As shown in Figure \ref{fig:weight-decay}, the weight decay used in the optimization has a significant effect on the learned weights. When the weight decay is active, the strength of weights will converge around the first 100 iterations of training. Further, the strength of frequencies is ordered with respect to the wave numbers where the lower frequencies have high strength and the higher frequencies have lower strength. On the other hand, when training without weight decay, the strength of all frequency modes will continuously grow and mix up with each other.

%% file: sections/appendix/additional_results.tex
\newpage
\section{Experiment Setup}

\paragraph{Burgers' Equation}
We consider the 1D Burgers' equation, which is a non-linear PDE with various applications, including modeling the flow of a viscous fluid. The 1D Burgers' equation takes the form:
$$
\begin{aligned}
\partial_t u(x, t)+\partial_x\left(u^2(x, t) / 2\right) & =\nu \partial_{x x} u(x, t), & & t \in(0,1] \\
u(x, 0) & =u_0(x), & & x \in(0,1)
\end{aligned}
$$
where $u_0$ is the initial condition and $\nu$ is the viscosity coefficient. We aim to learn the operator $\mathcal{G}_\theta$, mapping the initial condition to the solution. The training is performed under the resolution of 1024, and we consider the learning rate halved every 50 epochs and weight decay to be 0.0001. 

\paragraph{Darcy Flow}
Next, we consider the steady-state of the 2D Darcy Flow on the unit box, which is the second-order, linear, elliptic PDE:
$$
\begin{aligned}
-\nabla \cdot(a(x) \nabla u(x)) & =f(x), & & x \in(0,1)^2 \\
u(x) & =0, & & x \in \partial(0,1)^2
\end{aligned}
$$
With a Dirichlet boundary where $a$ is the diffusion coefficient, and $f=1$ is the forcing function. We are interested in learning the operator mapping the diffusion coefficient to the solution. The resolution is set to be $421 \times 421$.

\paragraph{Navier-Stokes Equation}
\input{figures/wrap_main_results.tex}
We also consider the 2D+time Navier-Stokes equation for a viscous, incompressible fluid on the unit torus:
$$
\begin{aligned}
\partial_t w(x, t)+u(x, t) \cdot \nabla w(x, t) & =\nu \Delta w(x, t)+f(x), \\
\nabla \cdot u(x, t) & =0,  \; \; x \in(0,1)^2 \\
w(x, 0) & =w_0(x), \; \; t \in(0, T]
\end{aligned}
$$
where $w=\nabla \times u$ is the vorticity, $w_0$ is the initial vorticity, $\nu \in \mathbb{R}_{+}$is the viscosity coefficient, and  $f(x) = 0.1\bigl( \mathrm{sin}(2\pi(x_1 + x_2))+\mathrm{cos}(2\pi(x_1 + x_2))\bigr)$ is the forcing function. We are interested in learning the operator mapping the vorticity up to time 10 to the vorticity up to some later time. We experiment with the hardest viscosity task, i.e., $\nu=1 \mathrm{e}-5$. The equivalent Reynolds number is around $500$ normalized by forcing and domain size. We set the final time $T=20$ as the dynamic becomes chaotic. 
The resolution is fixed to be $64 \times 64$.

We consider solving this equation using the 2D FNO with an RNN structure in time or using the 3D FNO to do space-time convolution, similar to the benchmark in \citet{li_fourier_2021}.
As 3D FNO requires significantly more parameters, evaluating 3D FNO with more than 30 modes is computationally intractable due to the hardware limitation. 

\subsection{Experiment details for Re5000 dataset - iFNO}
Regarding setting the scheduler, the incremental resolution approach significantly alters the loss landscape and distribution of the training data each time the resolution is increased. As noted by \cite{smith2017cyclical}, cyclic learning rates oscillating between higher and lower values during training can be highly beneficial when dealing with such changing loss landscapes. Therefore, our incremental resolution training methodology adopts a combination of the cyclic learning rate and step learning rate scheme. We see that this combination provides improved and more stable convergence throughout the curriculum of incremental resolution increases, contributing to strong empirical results for both the incremental resolution model and the iFNO model. A triangular cyclic learning rate allows the model to first explore with larger learning rates to find new optimal parameters after a resolution increase before narrowing down with smaller rates to converge. This prevents the model from getting stuck in sub-optimal solutions. Our experiments utilize 3 different modes, i.e., triangular, triangular with decreased height, and exponential decay. We also sweep over various step sizes and maximum and minimum bounds. The cyclic period is set to $e$ epochs. Although the cyclicLR method is indeed suitable, we find that after we switch to the highest resolution, using stepLR converges to a better minimum compared to cyclicLR. This makes sense, as once we start training on the highest resolution, instead of changing our learning rates in an oscillatory manner, we should stabilize and take smaller steps for convergence, like how the baseline FNO is trained.

Our algorithm increases resolution every $e$ epoch from 32 to 64 and then 128. We use powers of 2 for optimal Fast Fourier Transform speed. The hyperparameter $e$ can be tuned, and $r$ is a list determining how low we sub-sample data. In the re5000 dataset case, we have set $e = 10$ and $r = [4, 2, 1]$, which results in starting from resolution 32 all the way up to the highest resolution 128.

\section{Additional Results}
\label{sec:additional-results}

\subsection{Full Data Results}

We evaluate the models with the most data samples available in each dataset.
As shown in Table~\ref{tab:main_results_full_data}, iFNO achieves good performance comparable with standard FNO. It even achieves the best testing performance on Navier Stokes (2D) and Kolmogorov Flow.

\begin{table*}[!h]
\centering
  \caption{\small{Evaluation on various PDEs in the full data regime. The average training and testing L2 loss are reported with their standard deviation over 3 repeated runs. Frequency-based and loss-based iFNO are compared with standard FNO with fixed $K = 10, 30, 60, 90$ number of modes. Losses have been divided by the scale under the column. 
  }}
\begin{tabular}{lccccccc}
    \toprule
    \textbf{Method} &
      \multicolumn{2}{c}{\textbf{Burgers}} &
      \multicolumn{2}{c}{\textbf{Darcy Flow}}  &
      \multicolumn{2}{c}{\textbf{Navier Stokes (FNO 2D)}} \\
      & {Train (\num{1e-3})} & {Test (\num{1e-3})} & {Train (\num{1e-4})} & {Test (1e-4)} & {Train (\num{1e-1})} & {Test (1e-1)}\\
      \cmidrule(r){2-3}\cmidrule(r){4-5}\cmidrule(r){6-7}
        iFNO (Freq) & $0.432 \pm 0.018$ & {0.546} $\pm$ {0.020} & {0.447} $\pm$ {0.006} & {0.509} $\pm$ {0.020} & $1.383 \pm 0.015$ & \textbf{1.905} $\pm$ \textbf{0.017}\\
        iFNO (Loss) & $0.391 \pm 0.027$ & $ 0.595 \pm 0.105$ & {0.405} $\pm$ {0.034} & $0.464 \pm 0.048$ & \textbf{1.325} $\pm$ \textbf{0.003} & $1.918 \pm 0.006$\\
        \cmidrule(r){2-7}
        FNO (10) & $0.441 \pm 0.015$ & $0.576 \pm 0.006$ & {0.422} $\pm$ {0.012} & $0.479 \pm 0.019$ & $1.409 \pm 0.006$ & $1.912 \pm 0.021$\\
        FNO (30) & \textbf{0.382} $\pm$ \textbf{0.008} & \textbf{0.499} $\pm$ \textbf{0.003} & \textbf{0.390} $\pm$ \textbf{0.024} & \textbf{0.449} $\pm$ \textbf{0.026} & $1.346 \pm 0.011$ & $1.934 \pm 0.012$\\
        FNO (60) & $0.385 \pm 0.017$ & $0.518 \pm 0.004$ & {0.433} $\pm$ {0.015} & $0.485 \pm 0.020$ & {1.350} $\pm$ {0.011} & $1.931 \pm 0.012$\\
        FNO (90) & {0.394} $\pm$ {0.011} & $0.540 \pm 0.039$ & {0.424} $\pm$ {0.006} & $0.489 \pm 0.030$ & $1.335 \pm 0.012$ & $1.927 \pm 0.006$\\
    \bottomrule
  \end{tabular}
  \vskip 0.1in
\centering
\begin{tabular}{lcccc}
    \toprule
    \textbf{Method} &
      \multicolumn{2}{c}{\textbf{Navier Stokes (FNO 3D)}} &
     \multicolumn{2}{c}{\textbf{Kolmogorov Flow}}\\
      & {Train (\num{1e-2})} & {Test (\num{1e-1})} & {Train (\num{1e-2})} & {Test (\num{1e-2})}\\
      \cmidrule(r){2-3}\cmidrule(r){4-5}
        iFNO (Freq) & $4.843 \pm 0.116$ & $4.855 \pm 0.005$ & {3.271} $\pm$ {0.025} & \textbf{4.343} $\pm$ \textbf{0.033}\\
        iFNO (Loss) & $2.798 \pm 0.106$ & $4.879 \pm 0.013$ & {3.219} $\pm$ {0.139} & {5.032} $\pm$ {0.301}\\
        \cmidrule(r){2-5}
        FNO (10) & $5.596 \pm  0.030$ & \textbf{4.813} $\pm$ \textbf{0.013} & {12.26} $\pm$ {0.057} & $13.28 \pm 0.015$\\
        FNO (30) & \textbf{2.776} $\pm$  \textbf{0.098} & $4.877 \pm 0.008$ & {5.941} $\pm$ {0.023} & $6.775 \pm 0.019$\\
        FNO (60) & $-$ & $-$ & {3.789} $\pm$ {0.022} & $4.865 \pm 0.030$\\
        FNO (90) & $-$ & $-$ & \textbf{3.091} $\pm$ \textbf{0.022} & $4.597 \pm 0.070$\\
    \bottomrule
  \end{tabular}
  \label{tab:main_results_full_data}
\end{table*}

\newpage
\subsection{Frequency mode evolution during training}
\begin{figure*}[!h]
    \centering
    \includegraphics[width=1\textwidth]{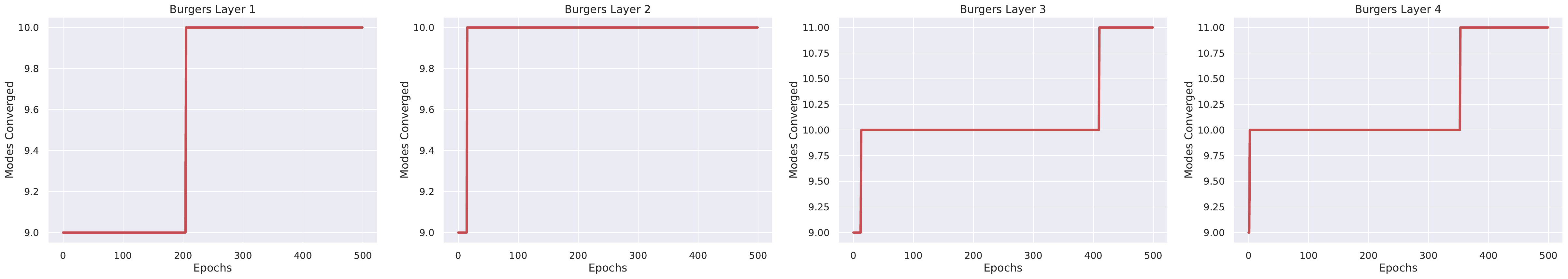}
    \includegraphics[width=1\textwidth]{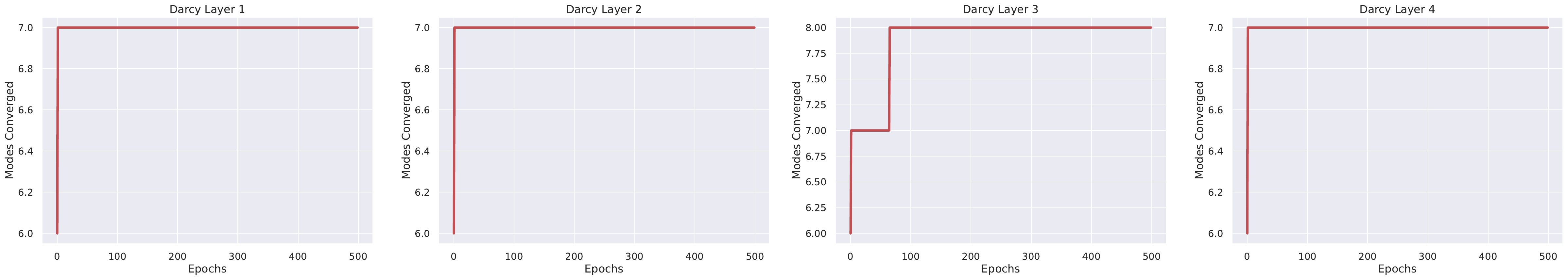}
    \includegraphics[width=1\textwidth]{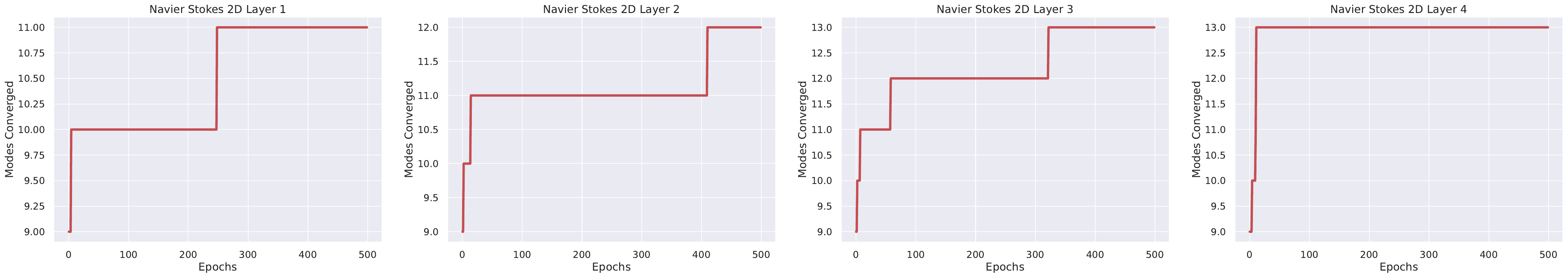}
    \includegraphics[width=1\textwidth]{figures/final-figures/modes_converged_Navier_Stokes_2D.pdf}
    \includegraphics[width=1\textwidth]{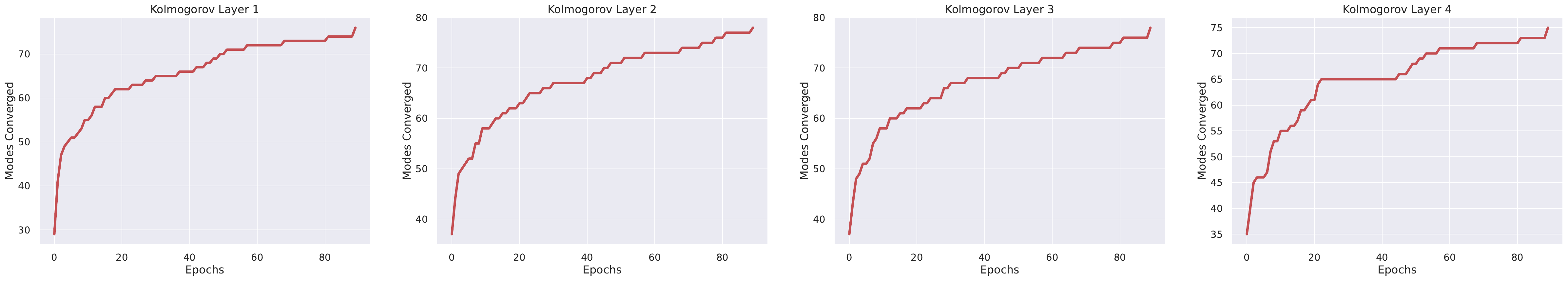}
    \caption{\small{Mode evolution during training for each layer across datasets.}}
    \label{fig:mode_plot}
\end{figure*}

As shown in Figure \ref{fig:mode_plot}, the effective modes $K$ adaptively increase during the training phase on the Burgers, Darcy, Navier-Stokes (2D \& 3D formulation), and Kolmogorov Flows.
For smooth problems such as viscous Burgers equation and Darcy Flow, the mode converges in the early epoch, while for problem with high frequency structures such as the Kolmogorov Flows, the mode will continuously increase throughout the training phase.

\newpage
\subsection{Training and testing loss during learning}
\begin{figure*}[!h]
    \centering
    \includegraphics[width=1\textwidth]{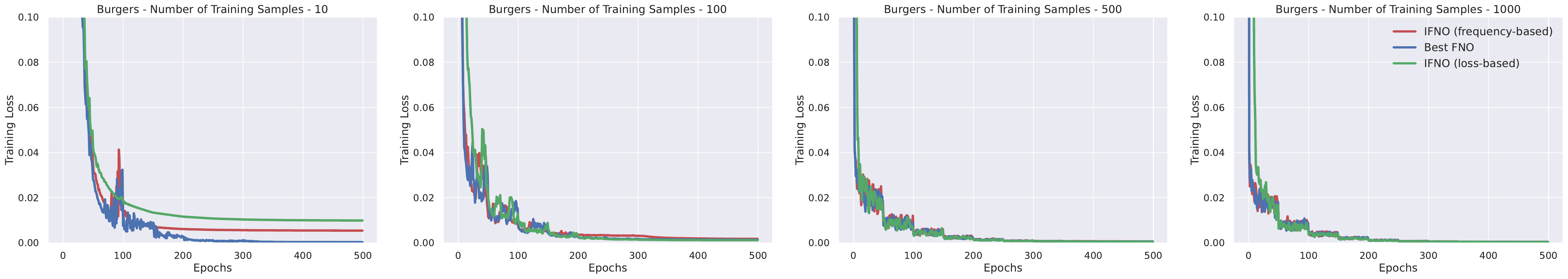}
    \includegraphics[width=1\textwidth]{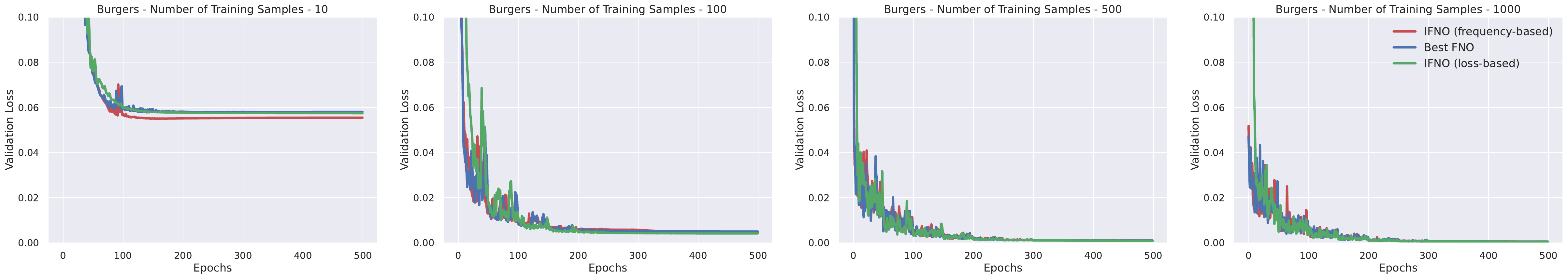}
    \includegraphics[width=1\textwidth]{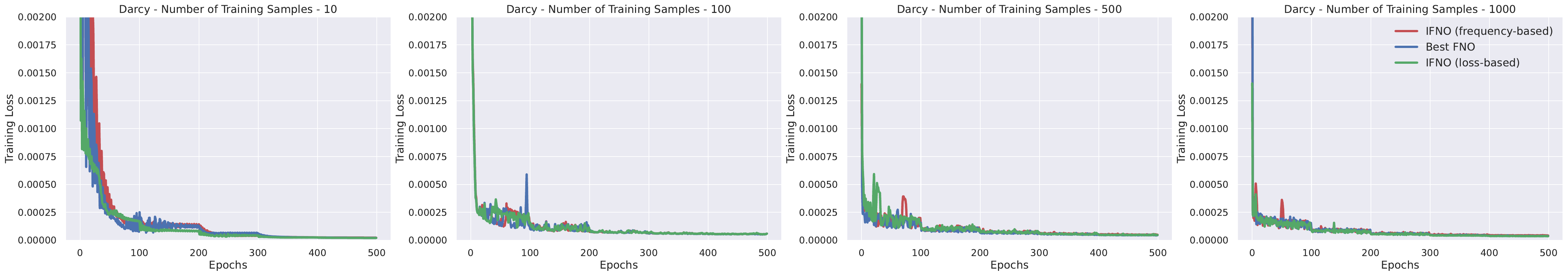}
    \includegraphics[width=1\textwidth]{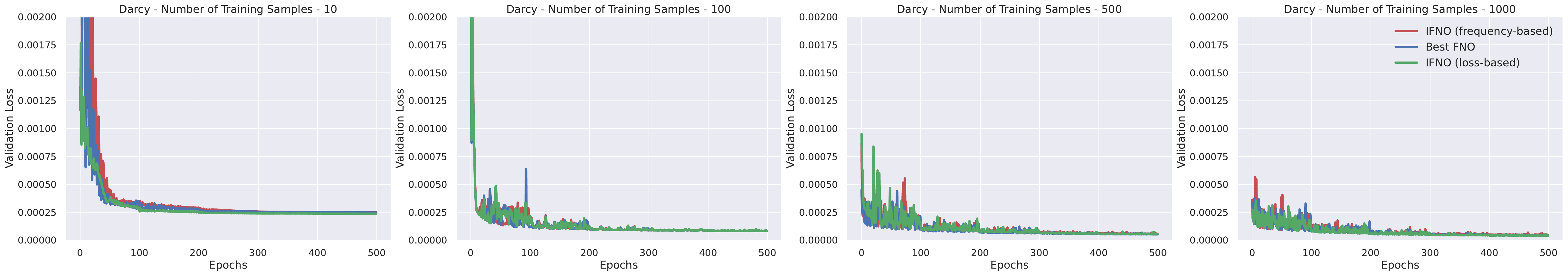}
    \includegraphics[width=1\textwidth]{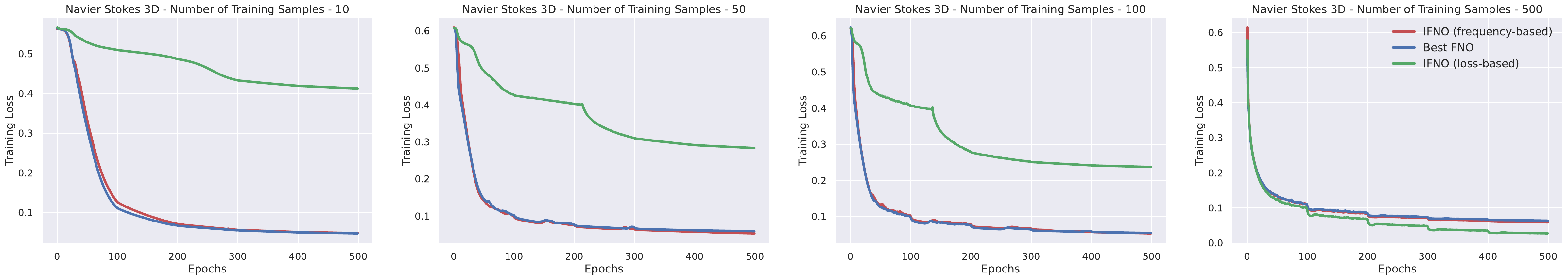}
    \includegraphics[width=1\textwidth]{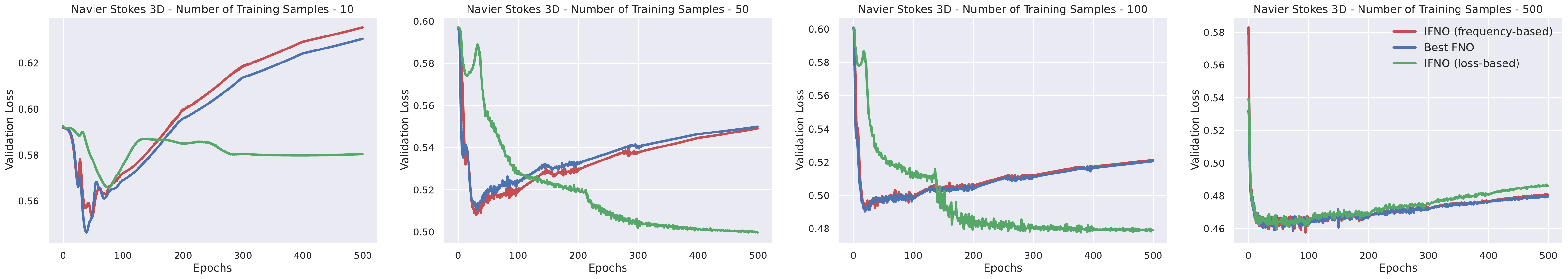}
    \caption{\small{Loss vs Epochs across datasets (part 1: Burgers, Darcy, Navier-Stokes 2D). \textbf{Each top row: training L2 loss}. \textbf{Each bottom row: validation L2 loss}}}
    \label{fig:loss_plot1}
\end{figure*}

\begin{figure*}[!h]
    \centering
    \includegraphics[width=1\textwidth]{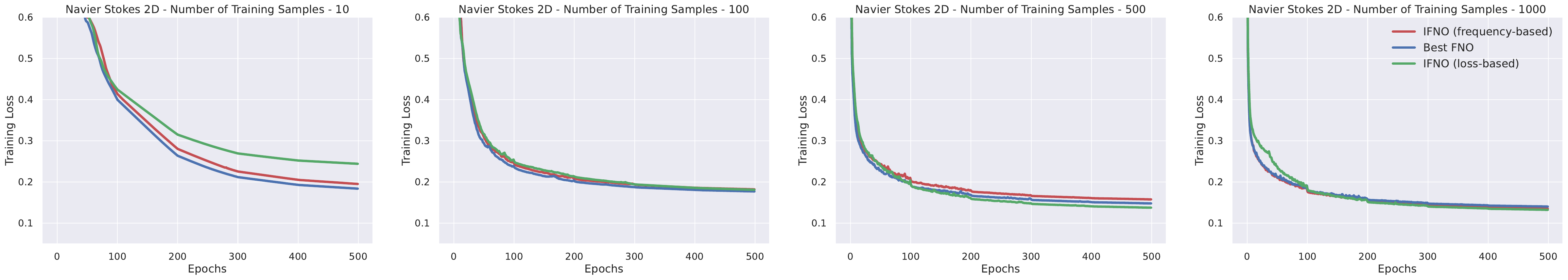}
    \includegraphics[width=1\textwidth]{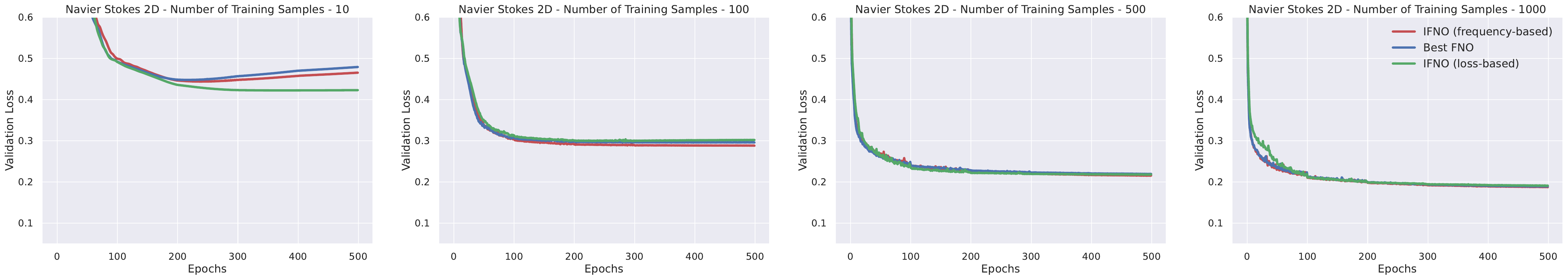}
    \includegraphics[width=1\textwidth]{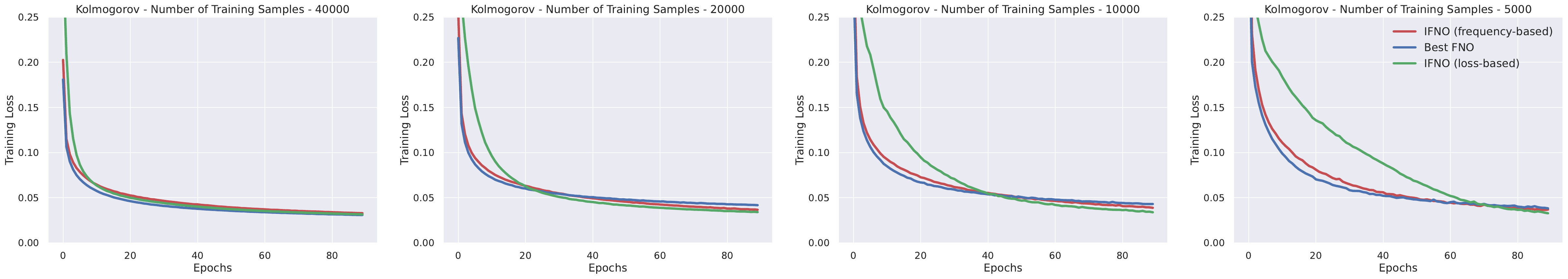}
    \includegraphics[width=1\textwidth]{figures/final-figures/lossKolmogorovtr.pdf}
    \caption{\small{Loss vs Epochs across datasets (part 2: Navier-Stokes 2D, Kolmogorov Flows). \textbf{Each top row: training L2 loss}. \textbf{Each bottom row: validation L2 loss}}}
    \label{fig:loss_plot2}
\end{figure*}

Figure \ref{fig:loss_plot1} and \ref{fig:loss_plot2} show the training curves and testing curves across the five problem settings, ten rows in total. 
Overall, iFNO methods have a slightly higher training error but lower validation error, indicating its better generalization over the few data regime.

%% file: sections/appendix/rebuttal_results.tex
\section{Additional Experiments and Results}

\subsection{Selection of Hyperparameters}

\begin{table*}[!h]
    \centering
      \caption{\small{Evaluation of $\alpha$ in iFNO (Freq) across various tasks.}}
    \begin{tabular}{lcccccc}
        \toprule
        \textbf{Tasks} &
          \multicolumn{5}{c}{\textbf{Selection of $\alpha$}} \\
          & 0.6 & 0.8 & 0.9 & 0.99 & 0.999 & 0.9999  \\
          \cmidrule(r){2-7} %
          {\textbf{Burgers} (1e-3)} & 1.723 & 0.975 & 0.583 & 0.544 & 0.559 & 0.587 \\
        {\textbf{Darcy Flow} (1e-4)} & 0.621 & 0.595 & 0.552 & 0.509 & 0.475 & 0.513 \\
        {\textbf{Navier Stokes (FNO 2D)}} & 0.341 & 0.256 & 0.223 & 0.189 & 0.194 & 2.010  \\
        {\textbf{Navier Stokes (FNO 3D)}} & 0.524 & 0.499 & 0.461 & 0.464 & 0.468 & 0.471 \\
        {\textbf{Kolmogorov Flow}} & 0.133 & 0.122 & 0.102 & 0.043 & 0.049 & 0.051  \\
        \bottomrule
      \end{tabular}
      \vskip 0.1in
      \label{tab:selection_alpha}
    \end{table*}

\pagebreak

\begin{table*}[!h]
    \centering
      \caption{\small{Evaluation of $\eta$ in iFNO (Loss) across various tasks.}
      }
    \begin{tabular}{lccccccc}
        \toprule
        \textbf{Tasks} &
          \multicolumn{5}{c}{\textbf{Selection of $\eta$}} \\
          & 0.1 & 0.01 & 0.001 & 0.0001 & 0.00001 \\
          \cmidrule(r){2-6}
          {\textbf{Burgers} (1e-3)} & 0.672 & 0.572 & 0.595 & 1.190 & 13.90 \\
        {\textbf{Darcy Flow} (1e-4)} & 0.486 & 0.465 & 0.412 & 0.424 & 0.490 \\
        {\textbf{Navier Stokes (FNO 2D)}} & 0.192 & 0.194 & 0.191 & 0.284 & 0.322\\
        {\textbf{Navier Stokes (FNO 3D)}} & 0.471 & 0.469 & 0.454 & 0.449 & 0.452 \\
        {\textbf{Kolmogorov Flow}} & 0.0614 & 0.0616 & 0.053 & 0.578 & 0.579 \\
        \bottomrule
      \end{tabular}
      \vskip 0.1in
      \label{tab:selection_eta}
    \end{table*}

\subsection{FNO over different sizes}

\begin{table*}[!h]
    \centering
      \caption{\small{Evaluation of changing the width of FNO models on Burger. Validation losses have been divided by 1e-3.}
      }
    \begin{tabular}{lccccccc}
        \toprule
        \textbf{Models} &
          \multicolumn{5}{c}{\textbf{Dimension}} \\
          & 32 & 64 & 80 & 96 & 128 \\
          \cmidrule(r){2-6}
          {\textbf{iFNO (Freq)}}& 0.785 & 0.5901 & 0.532 & \textbf{0.469} & \textbf{0.513}\\
        {\textbf{iFNO (Loss)}} & 0.746 & 0.708 & 0.605 & 0.578 & 0.714\\
        {\textbf{FNO}} & \textbf{0.603} & \textbf{0.519} & \textbf{0.525} & 0.501 & 0.553\\
        \bottomrule
      \end{tabular}
      \vskip 0.1in
      \label{tab:selection_frequency}
    \end{table*}

\begin{table*}[!h]
\centering
\caption{\small{Evaluation of changing the width of FNO models on Darcy. Validation losses have been divided by 1e-4.}
}
\begin{tabular}{lccccccc}
\toprule
\textbf{Models} &
\multicolumn{5}{c}{\textbf{Dimension}} \\
& 32 & 64 & 80 & 96 & 112 \\
\cmidrule(r){2-6}
{\textbf{iFNO (Freq) }}& {0.462} & {0.462} & 0.4395 & {0.4304} & 0.4479\\
{\textbf{iFNO (Loss)}} & \textbf{0.4549} & \textbf{0.4553} & 0.4599 & \textbf{0.3859} & \textbf{0.4321}\\
{\textbf{FNO}} & 0.4683 & 0.5051 & \textbf{0.4116} & 0.4922 & {0.4834}\\
\bottomrule
\end{tabular}
\vskip 0.1in
\label{tab:selection_frequency}
\end{table*}

\begin{table*}[!h]
    \centering
      \caption{\small{Evaluation of changing the width of FNO models on Navier Stokes 2D. Validation losses have been divided by 1.}}
    \begin{tabular}{lccccc}
        \toprule
        \textbf{Models} &
          \multicolumn{4}{c}{\textbf{Dimension}} \\
          & 16 & 20 & 32 & 64 & 80 \\
          \cmidrule(r){2-6}
          {\textbf{iFNO (Freq) }} & 0.205 & 0.192 & \textbf{0.180} & \textbf{0.170} & \textbf{0.176} \\
        {\textbf{iFNO (Loss)}} & \textbf{0.199} & \textbf{0.191} & 0.181 & 0.177 & 0.177 \\
        {\textbf{FNO}} & {0.203} & {0.198} & {0.182} & 0.182 & 0.177 \\
        \bottomrule
      \end{tabular}
      \vskip 0.1in
      \label{tab:selection_frequency}
    \end{table*}

\begin{table*}[!h]
    \centering
      \caption{\small{Evaluation of changing the width of FNO models on Kolmogorov Flow. Validation losses have been divided by 1e-1.}}
    \begin{tabular}{lccccc}
        \toprule
        \textbf{Tasks} &
          \multicolumn{4}{c}{\textbf{Dimension}} \\
          & 16 & 20 & 32 & 64 & 80 \\
          \cmidrule(r){2-6}
          {\textbf{iFNO (Freq) }} & 1.513 & 1.450 & 1.312 & 0.634 & (NR)\\
        {\textbf{iFNO (Loss)}} & \textbf{0.879} & \textbf{0.821} & \textbf{0.654} & \textbf{0.534} & (NR)\\
        {\textbf{FNO}} & 1.072 & 1.024 & {0.872} & 0.743 & 0.721\\
        \bottomrule
      \end{tabular}
      \vskip 0.1in
      \label{tab:selection_frequency}
    \end{table*}
\pagebreak
\subsection{Incremental FNO under different initialization scales}

\begin{table*}[!h]
    \centering
      \caption{\small{Evaluation of different initialization scales of FNO/iFNO models on Kolmogorov Flow. We again do a full sweep and report the best values.}
      }
    \begin{tabular}{lccccccc}
        \toprule
        \textbf{Tasks} &
          \multicolumn{5}{c}{\textbf{Scale of Initialization}} \\
          & 1e-2 & 1e-1 & 1 & 1e1 & 1e2 \\
          \cmidrule(r){2-6}
          {\textbf{iFNO (Freq) }} & 0.064 & 0.062 & 0.061 & 0.064 & 0.067\\
        {\textbf{iFNO (Loss)}} & \textbf{0.050} & \textbf{0.054} & \textbf{0.070} & \textbf{0.053} & \textbf{0.055}\\
        {\textbf{FNO}} & 0.073 & 0.073 & 0.072 & 0.072 & 0.073\\
        \bottomrule
      \end{tabular}
      \vskip 0.1in
      \label{tab:selection_frequency}
\end{table*}

\subsection{Standard FNO with different frequency modes}

\begin{table*}[!h]
    \centering
      \caption{\small{Evaluation of different frequency modes on standard FNO on Kolmogorov Flow. In this task, the minimum value is bolded across each row. Validation losses have been divided by the value next to the dataset across each row.}
      }
    \begin{tabular}{lcccccccccc}
        \toprule
        \textbf{Tasks} &
          \multicolumn{9}{c}{\textbf{Frequency Modes}} \\
          & 10 & 20 & 30 & 40 & 60 & 80 & 90 & 100 & 120\\
          \cmidrule(r){2-10}
        {\textbf{Kolmogorov Flow (1e-1)}} & 1.327 & 0.906 & 0.674 & 0.598 & 0.482 & 0.509 & \textbf{0.451} & {0.471} & 0.485 \\
        \bottomrule
      \end{tabular}
      \vskip 0.1in
      \label{tab:selection_frequency}
    \end{table*}

\subsection{Number of Paramaters}

\begin{table*}[!h]
    \centering
      \caption{\small{Evaluation and number of parameters (in Millions) for the various models in Table \ref{tab:main_results_full_data}.}
      }
\begin{tabular}{lcccc}
    \toprule
    \textbf{Method} &
      \multicolumn{2}{c}{\textbf{Navier Stokes (FNO 2D)}}&
     \multicolumn{2}{c}{\textbf{Kolmogorov Flow}}\\
      & {Number of Parameters (Millions)} & {Test (\num{1e-1})} & {Number of Parameters (Millions)} & {Test (\num{1e-2})}\\
      \cmidrule(r){2-3}\cmidrule(r){4-5}
        iFNO (Freq) & 0.92M & \textbf{1.872} & 321M & \textbf{4.602}  \\
        iFNO (Loss) & 23.8M & 1.911 & 530M & 5.127\\
        \cmidrule(r){2-5}
        FNO (10) & \textbf{0.64M} & 1.941 & \textbf{6.57M} & 13.27 \\
        FNO (30) & 5.76M & 1.924 & 59.0M & 6.462 \\
        FNO (60) & 23.0M & 1.918 & 235M & 4.825\\
        FNO (90) & 51.8M & 1.918 & 530M & 4.596\\
    \bottomrule
  \end{tabular}
  \label{tab:paramaters}
\end{table*}  

\subsection{Tensorized Factorized FNO under different tensor ranks}
\label{tfno}
\begin{table*}[!h]
    \centering
      \caption{\small{Evaluation of 2 different factorization ranks of iTFNO models on Re5000. We do a full sweep and report the best values.}
      }
    \begin{tabular}{lccccccc}
        \toprule
        \textbf{Weight Factorization} &
          \multicolumn{2}{c}{\textbf{5\%}} & \multicolumn{2}{c}{\textbf{25\%}}\\
          & Test (1e-1) & Runtime (Mins) & Test (1e-1) & Runtime (Mins)\\
          \cmidrule(r){0-2} \cmidrule(r){3-5}
          {\textbf{iTFNO}} & \textbf{0.646} & \textbf{122} & \textbf{0.793} & \textbf{259}\\
          {\textbf{iTFNO (Freq) }} & 0.752 & 144 & 0.904 & 333\\
        {\textbf{TFNO}} & 0.719 & 166 & 0.872 & 280\\
        \bottomrule
      \end{tabular}
      \vskip 0.1in
      \label{tab:selection_frequency}
\end{table*}

%% file: main.bbl
\begin{thebibliography}{36}
\providecommand{\natexlab}[1]{#1}
\providecommand{\url}[1]{\texttt{#1}}
\expandafter\ifx\csname urlstyle\endcsname\relax
  \providecommand{\doi}[1]{doi: #1}\else
  \providecommand{\doi}{doi: \begingroup \urlstyle{rm}\Url}\fi

\bibitem[Basri et~al.()Basri, Jacobs, Kasten, and
  Kritchman]{basri_convergence_2019}
Basri, R., Jacobs, D., Kasten, Y., and Kritchman, S.
\newblock The convergence rate of neural networks for learned functions of
  different frequencies.
\newblock URL \url{http://arxiv.org/abs/1906.00425}.
\newblock type: article.

\bibitem[Bhatnagar et~al.(2019)Bhatnagar, Afshar, Pan, Duraisamy, and
  Kaushik]{bhatnagar2019prediction}
Bhatnagar, S., Afshar, Y., Pan, S., Duraisamy, K., and Kaushik, S.
\newblock Prediction of aerodynamic flow fields using convolutional neural
  networks.
\newblock \emph{Computational Mechanics}, 64\penalty0 (2):\penalty0 525--545,
  2019.

\bibitem[Boffetta et~al.(2012)Boffetta, Ecke, et~al.]{boffetta2012two}
Boffetta, G., Ecke, R.~E., et~al.
\newblock Two-dimensional turbulence.
\newblock \emph{Annual review of fluid mechanics}, 44\penalty0 (1):\penalty0
  427--451, 2012.

\bibitem[Cao et~al.()Cao, Fang, Wu, Zhou, and Gu]{cao_towards_2020}
Cao, Y., Fang, Z., Wu, Y., Zhou, D.-X., and Gu, Q.
\newblock Towards understanding the spectral bias of deep learning.
\newblock URL \url{http://arxiv.org/abs/1912.01198}.
\newblock type: article.

\bibitem[Dissanayake \& Phan-Thien(1994)Dissanayake and
  Phan-Thien]{dissanayake1994neural}
Dissanayake, M. and Phan-Thien, N.
\newblock Neural-network-based approximations for solving partial differential
  equations.
\newblock \emph{communications in Numerical Methods in Engineering},
  10\penalty0 (3):\penalty0 195--201, 1994.

\bibitem[Fridovich-Keil et~al.()Fridovich-Keil, Gontijo-Lopes, and
  Roelofs]{fridovich-keil_spectral_2022}
Fridovich-Keil, S., Gontijo-Lopes, R., and Roelofs, R.
\newblock Spectral bias in practice: The role of function frequency in
  generalization.
\newblock URL \url{http://arxiv.org/abs/2110.02424}.
\newblock type: article.

\bibitem[Grady~II et~al.(2022)Grady~II, Khan, Louboutin, Yin, Witte, Chandra,
  Hewett, and Herrmann]{grady2022towards}
Grady~II, T.~J., Khan, R., Louboutin, M., Yin, Z., Witte, P.~A., Chandra, R.,
  Hewett, R.~J., and Herrmann, F.~J.
\newblock Towards large-scale learned solvers for parametric pdes with
  model-parallel fourier neural operators.
\newblock \emph{arXiv preprint arXiv:2204.01205}, 2022.

\bibitem[Greenfeld et~al.(2019)Greenfeld, Galun, Basri, Yavneh, and
  Kimmel]{greenfeld2019learning}
Greenfeld, D., Galun, M., Basri, R., Yavneh, I., and Kimmel, R.
\newblock Learning to optimize multigrid pde solvers.
\newblock In \emph{International Conference on Machine Learning}, pp.\
  2415--2423. PMLR, 2019.

\bibitem[Guo et~al.(2016)Guo, Li, and Iorio]{guo2016convolutional}
Guo, X., Li, W., and Iorio, F.
\newblock Convolutional neural networks for steady flow approximation.
\newblock In \emph{Proceedings of the 22nd ACM SIGKDD international conference
  on knowledge discovery and data mining}, pp.\  481--490, 2016.

\bibitem[Hou \& Li(2007)Hou and Li]{hou2007computing}
Hou, T.~Y. and Li, R.
\newblock Computing nearly singular solutions using pseudo-spectral methods.
\newblock \emph{Journal of Computational Physics}, 226\penalty0 (1):\penalty0
  379--397, 2007.

\bibitem[Huang \& Alkhalifah()Huang and Alkhalifah]{huang_pinnup_2021}
Huang, X. and Alkhalifah, T.
\newblock {PINNup}: Robust neural network wavefield solutions using frequency
  upscaling and neuron splitting.
\newblock URL \url{http://arxiv.org/abs/2109.14536}.

\bibitem[Kissas et~al.(2022)Kissas, Seidman, Guilhoto, Preciado, Pappas, and
  Perdikaris]{kissas2022learning}
Kissas, G., Seidman, J.~H., Guilhoto, L.~F., Preciado, V.~M., Pappas, G.~J.,
  and Perdikaris, P.
\newblock Learning operators with coupled attention.
\newblock \emph{Journal of Machine Learning Research}, 23\penalty0
  (215):\penalty0 1--63, 2022.

\bibitem[Kochkov et~al.(2021)Kochkov, Smith, Alieva, Wang, Brenner, and
  Hoyer]{kochkov2021machine}
Kochkov, D., Smith, J.~A., Alieva, A., Wang, Q., Brenner, M.~P., and Hoyer, S.
\newblock Machine learning accelerated computational fluid dynamics.
\newblock \emph{arXiv preprint arXiv:2102.01010}, 2021.

\bibitem[Kossaifi et~al.(2023)Kossaifi, Kovachki, Azizzadenesheli, and
  Anandkumar]{kossaifi2023multi}
Kossaifi, J., Kovachki, N., Azizzadenesheli, K., and Anandkumar, A.
\newblock Multi-grid tensorized fourier neural operator for high-resolution
  pdes.
\newblock \emph{arXiv preprint arXiv:2310.00120}, 2023.

\bibitem[Kovachki et~al.(2021)Kovachki, Li, Liu, Azizzadenesheli, Bhattacharya,
  Stuart, and Anandkumar]{kovachki2021neural}
Kovachki, N., Li, Z., Liu, B., Azizzadenesheli, K., Bhattacharya, K., Stuart,
  A., and Anandkumar, A.
\newblock Neural operator: Learning maps between function spaces.
\newblock \emph{arXiv preprint arXiv:2108.08481}, 2021.

\bibitem[Krishnapriyan et~al.(2021)Krishnapriyan, Gholami, Zhe, Kirby, and
  Mahoney]{characterize_failure_modes}
Krishnapriyan, A., Gholami, A., Zhe, S., Kirby, R., and Mahoney, M.~W.
\newblock Characterizing possible failure modes in physics-informed neural
  networks.
\newblock In Ranzato, M., Beygelzimer, A., Dauphin, Y., Liang, P., and Vaughan,
  J.~W. (eds.), \emph{Advances in Neural Information Processing Systems},
  volume~34, pp.\  26548--26560. Curran Associates, Inc., 2021.
\newblock URL
  \url{https://proceedings.neurips.cc/paper/2021/file/df438e5206f31600e6ae4af72f2725f1-Paper.pdf}.

\bibitem[Lagaris et~al.(1998)Lagaris, Likas, and
  Fotiadis]{lagaris1998artificial}
Lagaris, I.~E., Likas, A., and Fotiadis, D.~I.
\newblock Artificial neural networks for solving ordinary and partial
  differential equations.
\newblock \emph{IEEE transactions on neural networks}, 9\penalty0 (5):\penalty0
  987--1000, 1998.

\bibitem[Li et~al.({\natexlab{a}})Li, Kovachki, Azizzadenesheli, Liu,
  Bhattacharya, Stuart, and Anandkumar]{li_fourier_2021}
Li, Z., Kovachki, N., Azizzadenesheli, K., Liu, B., Bhattacharya, K., Stuart,
  A., and Anandkumar, A.
\newblock Fourier neural operator for parametric partial differential
  equations, {\natexlab{a}}.
\newblock URL \url{http://arxiv.org/abs/2010.08895}.
\newblock type: article.

\bibitem[Li et~al.({\natexlab{b}})Li, Kovachki, Azizzadenesheli, Liu,
  Bhattacharya, Stuart, and Anandkumar]{li_neural_2020}
Li, Z., Kovachki, N., Azizzadenesheli, K., Liu, B., Bhattacharya, K., Stuart,
  A., and Anandkumar, A.
\newblock Neural operator: Graph kernel network for partial differential
  equations, {\natexlab{b}}.
\newblock URL \url{http://arxiv.org/abs/2003.03485}.

\bibitem[Li et~al.(2021)Li, Kovachki, Azizzadenesheli, Liu, Bhattacharya,
  Stuart, and Anandkumar]{li2021markov}
Li, Z., Kovachki, N., Azizzadenesheli, K., Liu, B., Bhattacharya, K., Stuart,
  A., and Anandkumar, A.
\newblock Markov neural operators for learning chaotic systems.
\newblock \emph{arXiv preprint arXiv:2106.06898}, 2021.

\bibitem[Liu et~al.(2022)Liu, Kovachki, Li, Azizzadenesheli, Anandkumar,
  Stuart, and Bhattacharya]{liu2022learning}
Liu, B., Kovachki, N., Li, Z., Azizzadenesheli, K., Anandkumar, A., Stuart,
  A.~M., and Bhattacharya, K.
\newblock A learning-based multiscale method and its application to inelastic
  impact problems.
\newblock \emph{Journal of the Mechanics and Physics of Solids}, 158:\penalty0
  104668, 2022.

\bibitem[Liu et~al.()Liu, Wu, and Wang]{liu_splitting_2019}
Liu, Q., Wu, L., and Wang, D.
\newblock Splitting steepest descent for growing neural architectures.
\newblock URL \url{http://arxiv.org/abs/1910.02366}.

\bibitem[Lu et~al.(2021)Lu, Jin, Pang, Zhang, and Karniadakis]{lu2021learning}
Lu, L., Jin, P., Pang, G., Zhang, Z., and Karniadakis, G.~E.
\newblock Learning nonlinear operators via deeponet based on the universal
  approximation theorem of operators.
\newblock \emph{Nature Machine Intelligence}, 3\penalty0 (3):\penalty0
  218--229, 2021.

\bibitem[Nelsen \& Stuart(2021)Nelsen and Stuart]{nelsen2021random}
Nelsen, N.~H. and Stuart, A.~M.
\newblock The random feature model for input-output maps between banach spaces.
\newblock \emph{SIAM Journal on Scientific Computing}, 43\penalty0
  (5):\penalty0 A3212--A3243, 2021.

\bibitem[Pathak et~al.(2021)Pathak, Mustafa, Kashinath, Motheau, Kurth, and
  Day]{pathak2021ml}
Pathak, J., Mustafa, M., Kashinath, K., Motheau, E., Kurth, T., and Day, M.
\newblock Ml-pde: A framework for a machine learning enhanced pde solver.
\newblock \emph{Bulletin of the American Physical Society}, 2021.

\bibitem[Pathak et~al.(2022)Pathak, Subramanian, Harrington, Raja,
  Chattopadhyay, Mardani, Kurth, Hall, Li, Azizzadenesheli,
  et~al.]{pathak2022fourcastnet}
Pathak, J., Subramanian, S., Harrington, P., Raja, S., Chattopadhyay, A.,
  Mardani, M., Kurth, T., Hall, D., Li, Z., Azizzadenesheli, K., et~al.
\newblock Fourcastnet: A global data-driven high-resolution weather model using
  adaptive fourier neural operators.
\newblock \emph{arXiv preprint arXiv:2202.11214}, 2022.

\bibitem[Rahaman et~al.()Rahaman, Baratin, Arpit, Draxler, Lin, Hamprecht,
  Bengio, and Courville]{rahaman_spectral_2019}
Rahaman, N., Baratin, A., Arpit, D., Draxler, F., Lin, M., Hamprecht, F.~A.,
  Bengio, Y., and Courville, A.
\newblock On the spectral bias of neural networks.
\newblock URL \url{http://arxiv.org/abs/1806.08734}.
\newblock type: article.

\bibitem[Raissi et~al.(2019)Raissi, Perdikaris, and
  Karniadakis]{raissi2019physics}
Raissi, M., Perdikaris, P., and Karniadakis, G.~E.
\newblock Physics-informed neural networks: A deep learning framework for
  solving forward and inverse problems involving nonlinear partial differential
  equations.
\newblock \emph{Journal of Computational Physics}, 378:\penalty0 686--707,
  2019.

\bibitem[Sirignano \& Spiliopoulos(2018)Sirignano and
  Spiliopoulos]{sirignano2018dgm}
Sirignano, J. and Spiliopoulos, K.
\newblock Dgm: A deep learning algorithm for solving partial differential
  equations.
\newblock \emph{Journal of computational physics}, 375:\penalty0 1339--1364,
  2018.

\bibitem[Smith(2017)]{smith2017cyclical}
Smith, L.~N.
\newblock Cyclical learning rates for training neural networks.
\newblock In \emph{2017 IEEE winter conference on applications of computer
  vision (WACV)}, pp.\  464--472. IEEE, 2017.

\bibitem[Soviany et~al.(2022)Soviany, Ionescu, Rota, and
  Sebe]{soviany2022curriculum}
Soviany, P., Ionescu, R.~T., Rota, P., and Sebe, N.
\newblock Curriculum learning: A survey.
\newblock \emph{International Journal of Computer Vision}, 130\penalty0
  (6):\penalty0 1526--1565, 2022.

\bibitem[Takamoto et~al.(2022)Takamoto, Praditia, Leiteritz, MacKinlay,
  Alesiani, Pfl{\"u}ger, and Niepert]{takamoto2022pdebench}
Takamoto, M., Praditia, T., Leiteritz, R., MacKinlay, D., Alesiani, F.,
  Pfl{\"u}ger, D., and Niepert, M.
\newblock Pdebench: An extensive benchmark for scientific machine learning.
\newblock \emph{arXiv preprint arXiv:2210.07182}, 2022.

\bibitem[Weinan \& Yu(2018)Weinan and Yu]{weinan2018deep}
Weinan, E. and Yu, B.
\newblock The deep ritz method: a deep learning-based numerical algorithm for
  solving variational problems.
\newblock \emph{Communications in Mathematics and Statistics}, 6\penalty0
  (1):\penalty0 1--12, 2018.

\bibitem[Wen et~al.(2022)Wen, Li, Azizzadenesheli, Anandkumar, and
  Benson]{wen2022u}
Wen, G., Li, Z., Azizzadenesheli, K., Anandkumar, A., and Benson, S.~M.
\newblock U-fno—an enhanced fourier neural operator-based deep-learning model
  for multiphase flow.
\newblock \emph{Advances in Water Resources}, 163:\penalty0 104180, 2022.

\bibitem[Xu et~al.(2019)Xu, Zhang, Luo, Xiao, and Ma]{xu2019frequency}
Xu, Z.-Q.~J., Zhang, Y., Luo, T., Xiao, Y., and Ma, Z.
\newblock Frequency principle: Fourier analysis sheds light on deep neural
  networks.
\newblock \emph{arXiv preprint arXiv:1901.06523}, 2019.

\bibitem[Zhu \& Zabaras(2018)Zhu and Zabaras]{zhu2018bayesian}
Zhu, Y. and Zabaras, N.
\newblock Bayesian deep convolutional encoder--decoder networks for surrogate
  modeling and uncertainty quantification.
\newblock \emph{Journal of Computational Physics}, 366:\penalty0 415--447,
  2018.

\end{thebibliography}
